%% file: main.tex
\documentclass[acmtog,nonacm]{acmart}

\setcopyright{none}
\acmVolume{}
\acmNumber{}
\acmArticle{}
\acmYear{}
\acmMonth{}
\acmDOI{}
\acmConference{}{}{}{}

\usepackage{booktabs} 

\usepackage[ruled]{algorithm2e} 

\SetAlFnt{\small}
\SetAlCapFnt{\small}
\SetAlCapNameFnt{\small}
\SetAlCapHSkip{0pt}

\begin{document}
\title{RoomCraft: Controllable and Complete 3D Indoor Scene Generation}

\author{Mengqi Zhou*}
\affiliation{%
 \institution{Institute of Automation, Chinese Academy of Sciences}
  \city{Beijing}
 \country{China}}
\email{zhoumengqi2022@ia.ac.cn}

\author{Xipeng Wang*}
\affiliation{%
 \institution{Institute of Automation, Chinese Academy of Sciences}
  \city{Beijing}
 \country{China}}
\email{wangxiping2024@ia.ac.cn}

\author{Yuxi Wang$\dagger$}
\affiliation{%
 \institution{Institute of Automation, Chinese Academy of Sciences}
  \city{Beijing}
 \country{China}}
\email{yuxiwang93@gmail.com}

\author{Zhaoxiang Zhang$\dagger$}
\affiliation{%
 \institution{Institute of Automation, Chinese Academy of Sciences}
  \city{Beijing}
 \country{China}}
\email{zhaoxiang.zhang@ia.ac.cn}

\fboxsep=0pt %
\fboxrule=0.4pt %

\begin{teaserfigure}
  \begin{center}
    \captionsetup[sub]{labelformat=parens}
    \includegraphics[width=.99\textwidth]{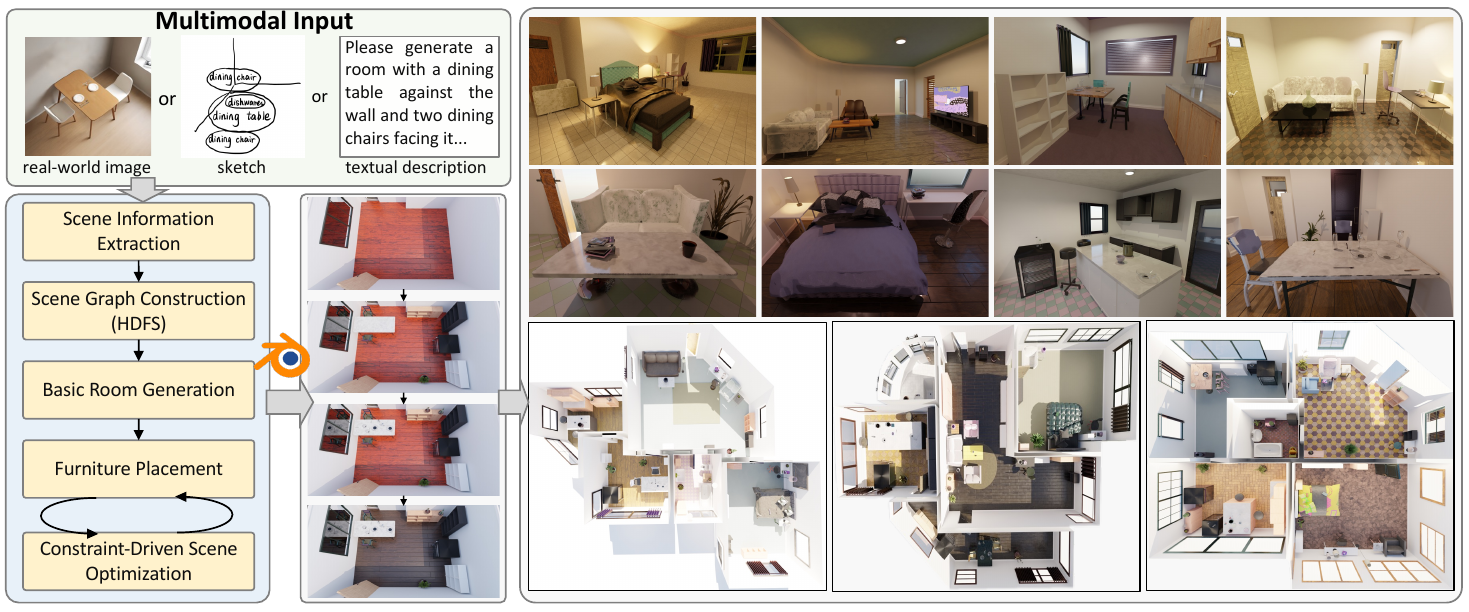}
    \caption{RoomCraft generates 3D room layouts based on user-provided inputs, including real-world images, sketches, or textual descriptions. The resulting scenes exhibit high diversity and closely align with the provided inputs, featuring realistic furniture arrangements and coherent spatial layouts.
    }
  \label{fig:teaser}
  \end{center}
\end{teaserfigure}

\input{sec/0_abstract}


\maketitle

\renewcommand{\thefootnote}{\fnsymbol{footnote}}
\footnotetext[0]{*Equal contribution.}
\footnotetext[0]{$\dagger$Corresponding authors.}

\renewcommand{\thefootnote}{\arabic{footnote}}

\input{sec/1_intro}
\input{sec/2_related_work}
\input{sec/3_method}

\input{sec/4_exper}

\input{sec/5_Conclusions}

\bibliographystyle{ACM-Reference-Format}
\bibliography{main}

\appendix

\input{sec/6_suppl}

\end{document}

%% file: sec/0_abstract.tex
\begin{abstract}
    Generating realistic 3D indoor scenes from user inputs remains a challenging problem in computer vision and graphics, requiring careful balance of geometric consistency, spatial relationships, and visual realism. While neural generation methods often produce repetitive elements due to limited global spatial reasoning, procedural approaches can leverage constraints for controllable generation but struggle with multi-constraint scenarios. When constraints become numerous, object collisions frequently occur, forcing the removal of furniture items and compromising layout completeness.
    To address these limitations, we propose RoomCraft, a multi-stage pipeline that converts real images, sketches, or text descriptions into coherent 3D indoor scenes. Our approach combines a scene generation pipeline with a constraint-driven optimization framework. The pipeline first extracts high-level scene information from user inputs and organizes it into a structured format containing room type, furniture items, and spatial relations. It then constructs a spatial relationship network to represent furniture arrangements and generates an optimized placement sequence using a heuristic-based depth-first search (HDFS) algorithm to ensure layout coherence. To handle complex multi-constraint scenarios, we introduce a unified constraint representation that processes both formal specifications and natural language inputs, enabling flexible constraint-oriented adjustments through a comprehensive action space design. Additionally, we propose a Conflict-Aware Positioning Strategy (CAPS) that dynamically adjusts placement weights to minimize furniture collisions and ensure layout completeness.
    Extensive experiments demonstrate that RoomCraft significantly outperforms existing methods in generating realistic, semantically coherent, and visually appealing room layouts across diverse input modalities.
\end{abstract}

%% file: sec/1_intro.tex
\section{Introduction}
\label{sec:intro}

In the field of Artificial General Intelligence (AGI), explicit generation \cite{SceneScape,hu2024scenecraft,SceneX,cityX} of 3D indoor scenes is essential for accurately simulating real-world environments, providing critical benefits such as geometric consistency, physical interactivity, and editability. These explicitly modeled scenes not only enable controlled generation and support physics-based rendering (PBR) but also offer reliable environments for robotics training and Embodied AI research.

However, generating plausible 3D indoor scenes remains challenging, especially in achieving realistic furniture arrangements within a space.
To address these problems, researchers commit to the implicit generation methods, including diffusion-based \cite{Text2Room,RoomDreamer,DiffuScene,InstructScene,EchoScene} and LLM-based methods \cite{Holodeck,LayoutGPT,AnyHome,Chat2Layout,LLplace}. Although these methods have shown promising results in 3D room layout generation, they may still struggle with repetitive elements and unrealistic layouts due to lacking global contextual understanding. On the other hand, explicit generation methods \cite{torne2024reconciling,Ditto,InfinigenIndoor,hu2024scenecraft,ProcTHOR} employ rule-based procedural generation to establish furniture positions and layouts. Due to the limitations of rule completeness and algorithmic constraints, these procedurally generated scenes frequently lack realism and fall short of providing the diversity needed for downstream tasks.

Alternatively, Digital Cousins \cite{ACDC} is an excellent work to extract high-level attributes from input images, such as semantic information, relative spatial relationships, and physical properties among furniture items, enabling the generation of accurate relative relationships. However, it overlooks the overall relationship between the furniture and the room structure, making it challenging to generate a complete room layout. InfineIndoor \cite{InfinigenIndoor} lacks visual priors and constraint information, resulting in poor controllability and visual harmony, exampling as out-of-boundary placement or inappropriate orientation. Therefore, we argue that beyond the \textit{diversity} and \textit{realism}, a 3D room generation method should also consider the following aspects: \\
\textit{\textbf{Controllability:} The ability to generate semantically and visually coherent room scenes based on a given instruction.} \\
\textit{\textbf{Completeness:} The ability to generate a well-organized, complete room layout based on a specific viewpoint input.}

In this paper, we introduce RoomCraft, a system designed to convert user-provided real images, sketches, or text descriptions into 3D indoor scenes. RoomCraft employs a multi-stage interactive generation pipeline, beginning with the extraction of high-level scene information from user inputs and its conversion into a structured, standardized format. This information includes room type, furniture categories, and spatial relationships between objects, leveraging the vision-language models (VLMs) to process various input modalities and ensure consistency across them. RoomCraft subsequently constructs a spatial relationship network to represent the relative positions of furniture items, generating a task list to guide furniture arrangement. The task list is organized through a heuristic-based depth-first search, prioritizing the placement of furniture items with stronger spatial constraints, thus ensuring layout coherence and spatial rationality. Additionally, to address complex multi-constraint scenarios in realistic room layouts, RoomCraft incorporates a constraint-driven optimization framework with three key components: a unified constraint representation system that processes both formal specifications and natural language inputs, an intelligent action space design that handles ambiguous spatial requirements, and a Conflict-Aware Positioning Strategy (CAPS) that prevents furniture collisions through dynamic weight adjustment. Unlike existing methods that compromise layout completeness when constraints become numerous, our approach maintains both constraint satisfaction and arrangement integrity, enabling the generation of semantically coherent and spatially rational 3D indoor scenes across diverse input modalities.

The contributions of this paper are summarized as follows:

\begin{enumerate}
    \item We propose a novel 3D indoor scene generation method, \textbf{RoomCraft}, that creates realistic 3D indoor scenes guided by the image, sketch, or text description.
    \item We introduce an adaptive layout optimization algorithm to resolve object conflicts and generate 3D indoor layouts that are spatially coherent and visually appealing.
    \item We have conducted extensive experiments to generate realistic 3D rooms, which perform better than existing state-of-the-art methods.
\end{enumerate}

%% file: sec/2_related_work.tex
\section{Related Works}
\noindent\textbf{Data-Driven Scene Generation.}
Early efforts~\cite{Henzler2019, NguyenPhuoc2019, Wu2016, Yang2019, Zhou2021} in 3D generation focused on learning the distribution of 3D shapes and textures from category-specific datasets. These methods evolved to enable more realistic scene generation using real-world data. For instance, Scan2BIM~\cite{Murali17} uses heuristics for wall detection to generate 2D floor plans, and Ochmann~\cite{Ochmann19} applies integer linear programming for layout inference. Furukawa~\cite{Cabral14} uses multi-view stereo to create 3D meshes of interiors. Recent advancements like the ACDC system~\cite{Dai24} introduce virtual environments that replicate real-world scenes' geometric and semantic features, improving generation efficiency and transfer for diverse robotic training environments.

\noindent\textbf{Procedural Based Scene Generation.}
Researchers have explored Procedural Content Generation (PCG) for both natural and urban 3D environments~\cite{ecosystems2022, rivers, lipp2011interactive, talton2011metropolis, Parcels, UrbanPattern2013}. Early works like PMC~\cite{PMC} focused on procedural city generation using 2D boundaries and mathematical algorithms, while more advanced methods, such as Infinigen~\cite{Infinigen}, enabled infinite generation of natural landscapes. However, these approaches lacked flexibility and customization. Recent efforts like 3DGPT~\cite{3D-GPT} introduced text-based control for more predictable scene generation, allowing users to provide high-level instructions. SceneX~\cite{SceneX} advances this by creating a PCGHub, where LLMs manage diverse asset generation and orchestrate complex 3D environments. By offering greater control and integrating a wider range of assets, SceneX enables the creation of richer, more adaptable scenes, marking a significant step towards scalable and customizable procedural generation systems.

\begin{figure*}
    \centering
    \includegraphics[width=1\textwidth]{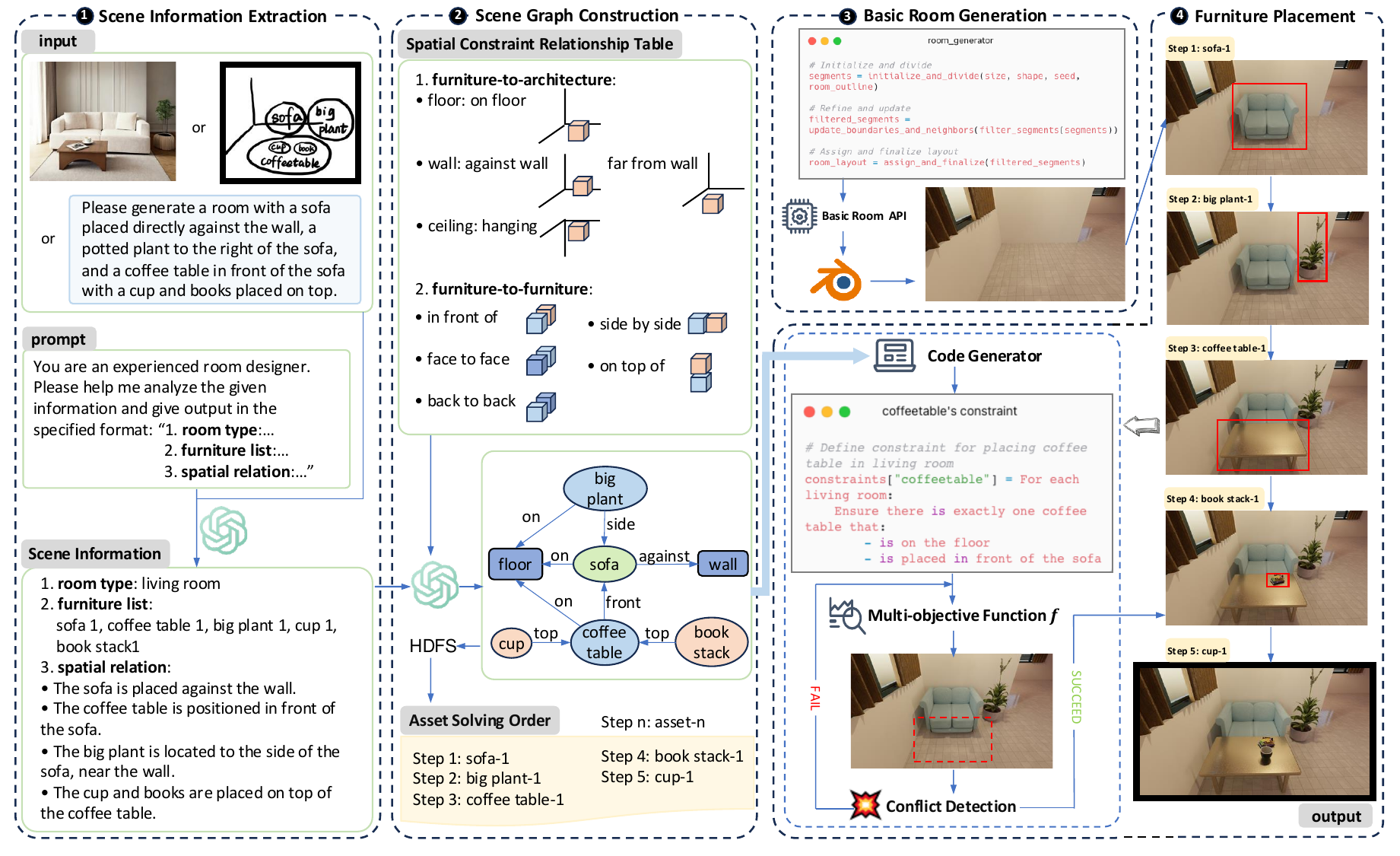}
    \vspace{-0.3in}
    \caption{RoomCraft: A Four-Stage Pipeline for Converting Multi-Modal Input into 3D indoor Scenes.}
    \label{fig:framework}
    \vspace{-0.4cm}
\end{figure*}

\noindent\textbf{Learning Based Scene Generation.}
Recent advancements in 3D scene generation, driven by machine learning and computer vision, have led to significant progress in generating both individual 3D objects~\cite{DreamFusion,Zero123,Magic3D,Fantasia3d} and expansive environments~\cite{Text2Room,Text2Nerf,SceneScape,SceneWiz3D,FurniScene}. Notably, methods like ZeroShot123~\cite{Zero123} employ diffusion models to generate 3D assets from images, while DreamFusion~\cite{DreamFusion} leverages NeRF to convert textual descriptions into 3D models. When it comes to large-scale scene generation, challenges increase, with systems like SceneDreamer~\cite{SceneDreamer} and CityDreamer~\cite{xie2023citydreamer} focusing on the creation of borderless, immersive environments. SceneDreamer uses BEV representations to generate interconnected 3D scenes, while CityDreamer models real-world city layouts. These methods represent a leap forward in creating detailed, expansive worlds, yet still face challenges in ensuring geometric clarity and compatibility with real-time engines like Unreal Engine, where issues like object intersection and spatial inconsistencies may occur.

%% file: sec/3_method.tex
\section{Method}

This section is organized into two parts: the scene generation pipeline of RoomCraft and the constraint-driven optimization framework. The first part presents the generation pipeline, which captures input layout—whether from text, real images, or hand-drawn sketches—and establishes spatial relationships between objects to create a coherent indoor environment. The second part introduces our constraint-driven optimization framework that systematically handles diverse spatial, functional, and aesthetic requirements through three key components: a unified constraint representation system that processes both formal specifications and natural language inputs, an intelligent action space design that enables flexible constraint-oriented adjustments, and a conflict-aware positioning strategy that proactively prevents spatial conflicts during furniture placement.

\subsection{A Multi-Stage Pipeline for Indoor Generation}

As shown in Fig. \ref{fig:framework}, our automated pipeline for 3D scene generation from textual descriptions, hand-drawn sketches, or real images follows four stages: (1) scene information extraction, where high-level room features and spatial relations are derived; (2) scene graph construction, where furniture items undergo constraint extraction and ordering based on spatial relationships; (3) basic room generation, defining essential elements like walls, doors, and windows; and (4) furniture placement, where items are sequentially arranged to create a coherent layout. 

\begin{figure*}[!tbp]
    \centering
    \includegraphics[width=0.99\linewidth]{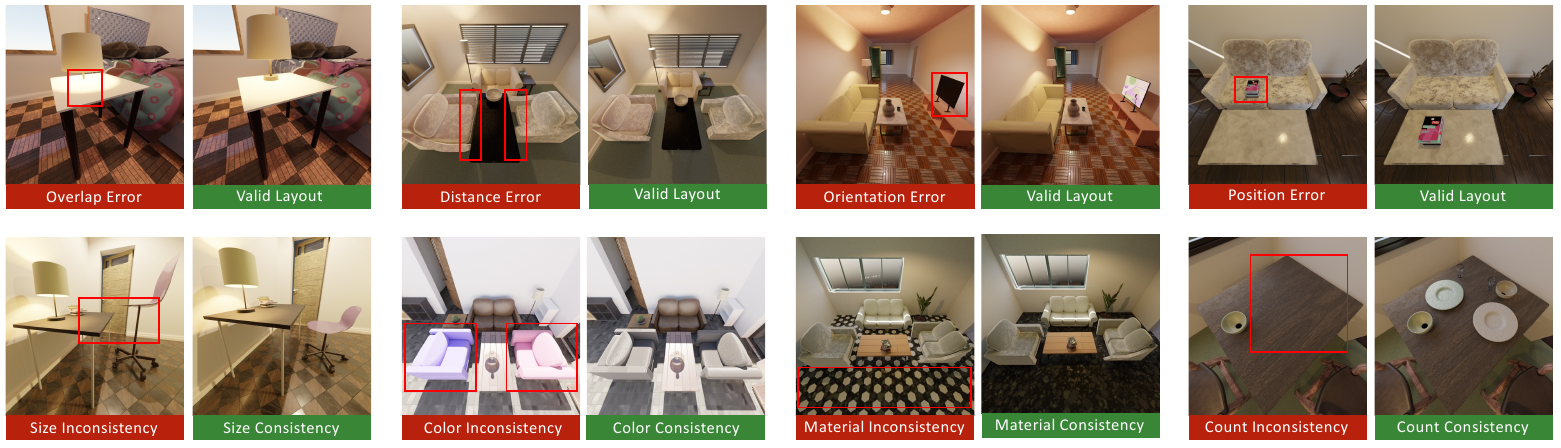}
    \caption{Constraint violation detection and correction results. Our framework identifies and resolves various constraint violations including spatial conflicts, attribute inconsistencies, and count errors. Each pair shows the detected violation (left, red boxes) and the corrected layout (right).}
    \label{fig:constraint_correction}
\end{figure*}

\noindent\textbf{Scene Information Extraction.} In the Scene Information Extraction stage of the RoomCraft system, the module extracts high-level information from input sources, including sketches, real-world images, or textual descriptions. Specifically, we use the VLM models, like GPT-4o, to process the input data \( I \) using corresponding prompts \( P_I \), producing a room scene organization $O$ as follows: 
\begin{equation}
    O = P_{GPT-4o}(I, P_I).
\end{equation}

Moreover, the organization $O$ is standardized into a JSON format, containing key fields such as \textit{RoomType ($R$)}, \textit{FurnitureItems ($V$)}, and \textit{SpatialRelations ($E$)}. 
This scene organization can be represented as:
\begin{equation}
    O = \left\langle R, V, E \right\rangle,
\end{equation}
where \( R \) denotes the room type, \( V \) represents the list of furniture items, and \( E \) is the set of spatial relations (\textit{e.g., in front of, on the top, etc.}). This structured output forms the foundation for the subsequent stage of the scene generation process, wherein the extracted data is used to generate a 3D representation of the room.

\noindent\textbf{Scene Graph Construction.} 
After obtaining the scene information, we provide a Scene Graph Construction method 
to extract and encode the spatial relationships between furniture items. 
Directly creating a complete room layout from descriptions is difficult due to the complexity and diversity of possible room arrangements. Inspired by previous works~\cite{hu2024scenecraft}, we first capture these relative spatial relationships, which can be more easily inferred, and use this information to guide the furniture arrangement in the room.

The extracted spatial relationships are represented as a set of constrained objects:
\begin{equation}
    \{\pi_{1,2}, \pi_{1,3}, ..., \pi_{m,n}\} \quad (m \neq n),
\end{equation}
where each object relationship \( \pi_{i,j} \) is defined as:
\begin{equation}
\pi_{i,j} = \langle V_i, V_j, E_{i,j} \rangle,
\end{equation}
where $V_i/V_j$ and $E_i$ are obtained from scene organization $O$, denoting a specific piece of furniture (\textit{e.g., table, chair, or sofa}), and the spatial constraints, respectively. Notably, these constraints capture relationships not only furniture-to-furniture (\textit{e.g., a cup on a table or a chair beside a table}) but also furniture-to-architecture (\textit{e.g., aligning a sofa with a wall}). These relationships are foundational guidelines for the layout generation stage, ensuring the coherent spatial relationship and orientation of furniture items.

\noindent\textbf{Heuristic-based Depth-First Search (HDFS). }
To achieve realistic and contextually appropriate room layouts, we design a heuristic-based depth-first search (HDFS) algorithm to provide rigorous constraints for the furniture and decorations.

The input of HDFS is the unordered set of constrained furniture objects, $\{V_i, V_2, ..., V_m\}$, and the output is an ordered arrangement $\{V_1^{'}, V_2^{'}, ..., V_m^{'}\} $, where each $V_i^{'}$ satisfies its respective constraints and follows a logical placement order.

Specifically, we first calculate a heuristic cost function \( f(V_i) \) for furniture \( V_i \) to prioritize objects with more critical constraints:
\begin{equation}
    f(V_i) = \sum_{j=1}^{m} w_i \cdot \mathbb{I}_j(V_i, V_j,E_{i,j}) (j\neq i),
\end{equation}
where \( w_i \) is the weight of constraint, indicating the importance of satisfying this constraint, and \( \mathbb{I}_i(V_i) \) is a binary indicator function.

Once the heuristic values \( f(V_i) \) are computed, objects are sorted in descending order based on these values to prioritise items with stronger constraints. The HDFS algorithm then iteratively places furniture according to this sorted list, progressively building the room layout while respecting spatial relationships. The final ordered layout can be formulated as: 
\begin{equation}
    \begin{split}
    {\{V_1^{'}, V_2^{'}, ..., V_m^{'}\}} = & \text{HDFS}(\{(V_1, f(V_1)), \\ &(V_2, f(v_2)),..., (V_m, f(V_m))\})        
    \end{split}
\end{equation}
where each \( V_i' \) represents the reordered furniture object that adheres to its specified spatial constraints, resulting in a coherent and logical room layout.

The HDFS algorithm ensures a logical furniture arrangement by prioritizing objects with stronger spatial constraints. For example, it places essential items, like a table, first, allowing subsequent objects, such as a cup, to be positioned correctly without further adjustments.

\begin{figure*}[!tbp]
    \centering
    \includegraphics[width=0.98\linewidth]{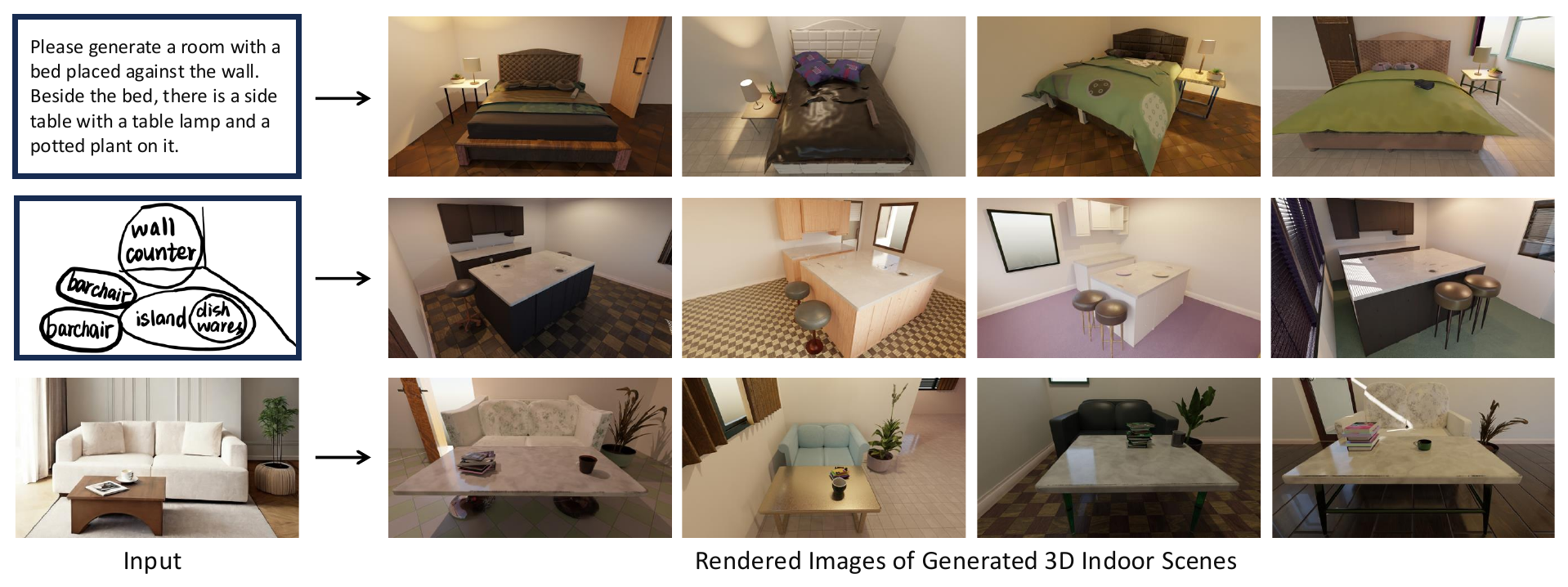}
    \caption{Results for instruct-to-room generation. We produce realistic 3D room scenes guided by three instruction types: textural description, sketch, and real-world RGB image. The generated rooms maintain consistency with input across room type, furniture categories, and placement locations.}
    \label{fig:instruct-to-room}
\end{figure*}

\noindent\textbf{Basic Room Generation.} 
Following Infinigen Indoors\cite{InfinigenIndoor}, we employ Procedural Content Generation software (\textit{e.g., Blender}) to create basic 3D room structures, leveraging its diverse APIs—covering semantic, relational, and scene generation functions—to facilitate the creation of basic room layouts while minimizing redundant tasks. The process begins with the procedural generation of essential spatial elements, including floors, walls, doors, and windows, which establish the spatial boundaries of the scene. This initial step creates the foundational structure required for further customization. Additionally, the system generates a room adjacency graph that defines the number, types, and connectivity of rooms. This graph guides the spatial arrangement of individual rooms, ensuring logical organization and connectivity between them. The resulting room structure offers a clear spatial framework, laying the groundwork for the subsequent placement of furniture and other scene elements.

\noindent\textbf{Furniture Placement.} 
To generate a coherent 3D layout, RoomCraft creates the entire scene using a single ``Addition" action to place furniture items sequentially. Unlike Infinigen Indoors, which employs a variety of discrete and continuous moves to edit objects, RoomCraft focuses on the efficient and systematic arrangement of furniture. The placement process begins with the object ordering provided by the Scene Graph Construction Stage $\{V_1^{'}, V_2^{'},...,V_n^{'}\}$. For each furniture item, an initial position is chosen on predefined surfaces, such as the floor or walls. This initial position is then adjusted based on spatial constraints to ensure alignment with the relational network and neighbouring objects. Through the ``Addition" action, RoomCraft establishes an organized layout that upholds spatial coherence within the room’s structural boundaries.

\begin{figure}[!tbp]
    \centering
    \includegraphics[width=0.95\linewidth]{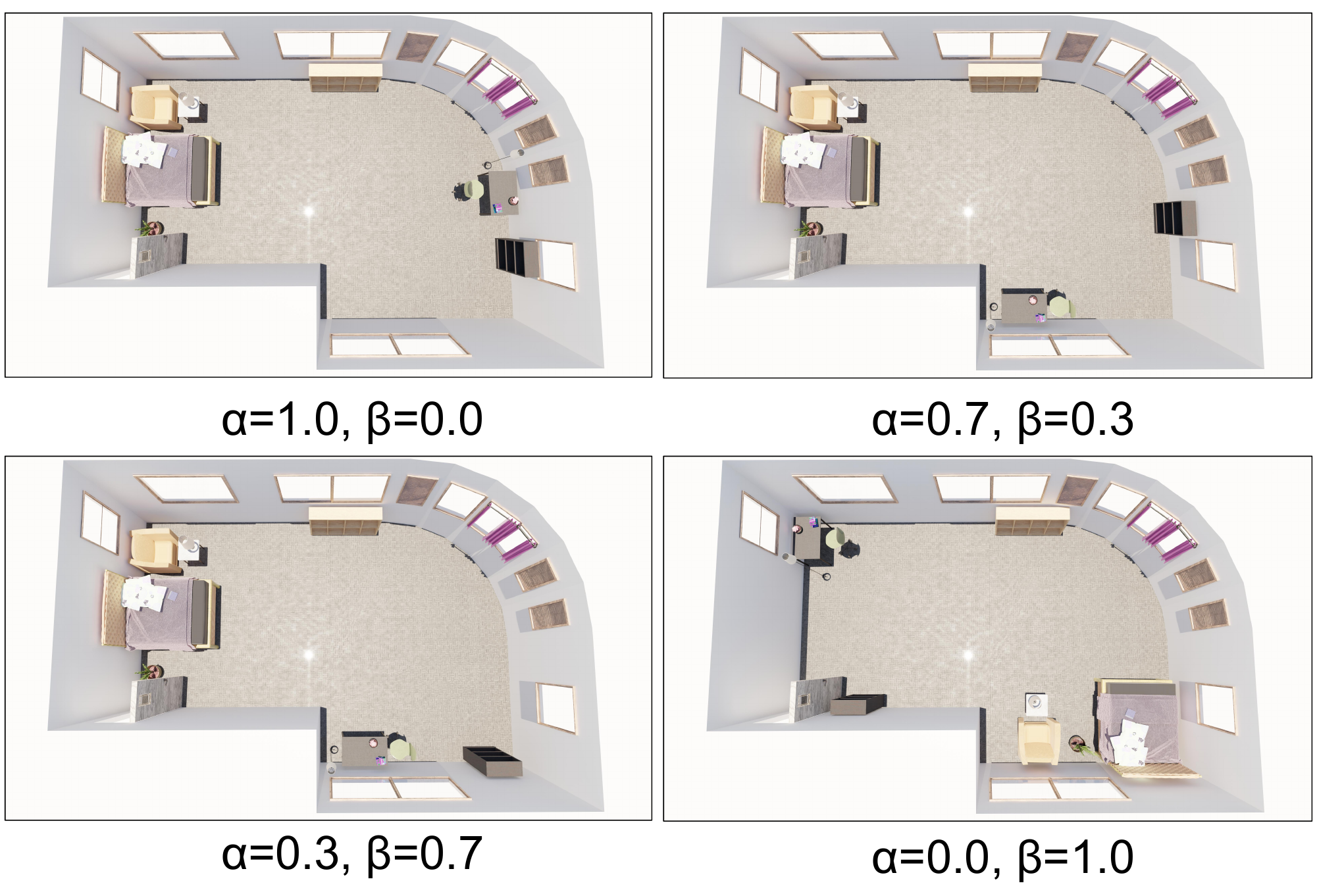}
    \vspace{-0.3cm}
    \caption{Influence of varying weights \( \alpha \) and \( \beta \) on spatial arrangement in room layout optimization.}
    \label{fig:alpha_beta}
    \vspace{-0.5cm}
\end{figure}

\subsection{Constraint-Driven Scene Optimization}

\noindent\textbf{Constraint Representation and Understanding.}
Indoor scene generation requires a unified constraint representation framework that can handle both formal and natural language inputs. We propose structured 5-tuples $C = (T, O, P, R, W)$, representing constraint type, involved objects, parameters, relationships, and weights respectively. This enables quantitative evaluation of spatial relationships, such as $\text{Distance}(\text{sofa}, \text{tv}) \in [2.0\text{m}, 3.5\text{m}]$. We distinguish between essential constraints that must be satisfied and flexible constraints that can be relaxed when necessary.
To automatically understand and process various constraint types, we design a collaborative framework combining LLMs, VLMs, and specialized constraint analysis tools. VLMs identify visual constraint violations from rendered scene images, while custom analysis functions detect geometric violations such as object overlaps through intersection volume calculations. LLMs then interpret these analysis results alongside textual constraint descriptions through structured prompt templates, generating adjustment instructions with specific actions, targets, parameters, and priorities.

\noindent\textbf{Action Space Design and Selection.}
We construct a constraint-oriented abstract action space that maps indoor design adjustment requirements to basic operations, encompassing physical adjustments (position, orientation, size), visual attributes (color, material), and high-level operations for object count and spatial organization. This abstraction enables LLMs to perform flexible reasoning and handle ambiguous constraints such as "the distance between the sofa and TV feels unnatural."
The LLM follows a multi-step decision process: first prioritizing constraints, then evaluating possible actions, and finally selecting an action sequence that balances constraint satisfaction and adjustment simplicity.

\noindent\textbf{Conflict-Aware Positioning Strategy (CAPS)}
In the furniture placement stage, collision risks increase with more furniture items, making suitable initial positioning crucial. We introduce a conflict-aware positioning strategy to address this challenge. Based on observations that accurately positioning wall-adjacent furniture significantly reduces collisions, we assume optimal placement should satisfy two conditions: sufficient distance from the opposite wall and minimal proximity to surrounding objects. We define a placement objective function to guide candidate position selection:
\begin{equation}
    \mathcal{L}_{plmt}(p) = \alpha \cdot \mathcal{L}_{dist}(p) + \beta \cdot \mathcal{L}_{obj}(p, \mu)
\end{equation}
where $\mathcal{L}_{dist}$ indicates the distance from position $p$ to the opposite wall, $\mathcal{L}_{obj}$ denotes the average distance from nearby objects within radius $\mu$ (set to 3), and $\alpha$, $\beta$ are controllable weights. To handle complex room layouts and conflicts, we employ dynamic weight adjustment where $\alpha$ increases for furniture-to-wall collisions and $\beta$ increases for furniture-to-furniture conflicts, maintaining the constraint $\alpha + \beta = 1$. The adjustment follows:
\begin{equation}
\alpha = \alpha + k \cdot \Delta_{\alpha}, \quad \beta = 1 - \alpha
\end{equation}
where $\Delta_{\alpha}$ represents the increment based on detected conflict type and $k$ controls the adjustment rate.


%% file: sec/4_exper.tex
\section{Experiments}

\subsection{Benchmark Protocol}

\textbf{Dataset.}
To investigate and evaluate the performance of RoomsCraft in indoor scene generation, we generate 150 room scenes constrained by 50 textual descriptions generated from GPT-4o\footnote{https://openai.com/index/hello-gpt-4o/}, 50 hand-drawn sketch images, and 50 RGB images obtained from the website.
Additionally, to facilitate comprehensive comparisons with various baselines, we manually modify the text descriptions to align with the input specifications of different baseline models.

\noindent\textbf{Baselines.} We compare our methods with the state-of-the-art approaches for 3D indoor scene generation. 
     \textit{InstructScene \cite{InstructScene}:} a diffusion-based method by integrating a semantic graph prior and a layout decoder. 
    \textit{LayoutGPT \cite{LayoutGPT}:} enhances LLMs with visual planning capabilities by generating layouts from text conditions. 
    \textit{AnyHome \cite{AnyHome}:} an 
    LLM-based methods that converting textual descriptions into indoor scenes. 
    \textit{Chat2Layout \cite{Chat2Layout}:} an interactive furniture layout generation system that enhances MLLMs with a unified VQ paradigm and a novel visual-text prompting mechanism. 
    \textit{Holodeck \cite{Holodeck}:} a LLM-based system that generates 3D environments to match a user-supplied prompt.
    \textit{Infinigen Indoor \cite{InfinigenIndoor}:} a rule-based procedural generation method that creates indoor 3D scenes from structured data representations.

\noindent\textbf{Metrics.}
    To evaluate the quality of spatial relation and orientation, we leverage two important metrics, including \textit{Out-of-Bound Rate (OOB)} and \textit{Orientation Correctness (ORI)} inspired Chat2Layout \cite{Chat2Layout}. The former assesses the spatial quality by measuring the percentage of layouts where objects either extend beyond the room boundaries or intersect unrealistically with other objects. The latter evaluates whether objects are oriented inaccessible and functional ways within the room. Moreover, to evaluate the overall quality of synthesized scenes with the guidance of text and image descriptions, we report \textit{CLIP Similarity (CLIP-Sim)} that assesses alignment between the user’s instructions and the scene content by calculating the CLIP similarity

    \begin{table}[!tbp]
    \centering
    \caption{Layout quality comparison. Our method achieves the best performance with a smaller Out-of-Boundary Rate (OOB), better Orientation Correction (ORI), and higher CLIP-Sim. $\dagger$ indicates the results are borrowed from Chat2Layout \cite{Chat2Layout}.}
    \label{tab:OOB}
    \begin{tabular}{lccc}
    \toprule
    \textbf{Method} & \textbf{OOB}$\downarrow$ & \textbf{ORI}$\uparrow$ & \textbf{CLIP-Sim} $\uparrow$ \\
    \midrule
    HOLODECK & 63.3 & 44.8 & 13.5 \\
    LayoutGPT & 52.8 & 61.3 & 18.4 \\
    InstructScene & 38.8 & 73.3 & 13.1 \\
    AnyHome & 32.8 & 72.3 & 23.2 \\
    Infinigen Indoors & 21.6 & 75.3 & - \\
    Chat2Layout & 21.0 & 84.8 & 27.1 \\
    \midrule
    \textbf{Ours (w/o CAPS)} & 20.7 & 87.3 & 27.4 \\
    \textbf{Ours} &  \textbf{19.3} & \textbf{89.3} & \textbf{27.7} \\
    \bottomrule
    \end{tabular}
\end{table}
\vspace{0.5cm}

\subsection{Experimental Results}

\subsubsection{Qualitative Results}

    \textbf{Instruct-to-room generation results.} We first provide  3D indoor scene generation results guided by three different types of instruction, including textual description, sketch, and real-world RGB images. The detailed results are depicted in Figure \ref{fig:instruct-to-room}. We provide front-view rendering images for different rooms. From the results, we observe that RoomCraft achieves alignment with the reference guidance across key aspects such as room type, furniture variety and placement. The promising outcomes further highlight the method's ability to accurately replicate spatial arrangements and aesthetic details, with the generated scenes closely mirroring the architectural nuances and stylistic elements of the reference input. These findings demonstrate the effectiveness of our approach in capturing complex spatial relationships. 
    Moreover, as we can create various different room according to one prompt instruction, it also reveals the diversity of our method.

    \noindent\textbf{Constraint-driven optimization results.} We demonstrate our constraint-driven optimization framework through comprehensive violation detection and correction examples. The detailed results are shown in Figure \ref{fig:constraint_correction}. We present before-and-after comparisons for various constraint violations, including spatial conflicts (overlap, distance, orientation, position errors), attribute inconsistencies (size, color, material mismatches), and object count discrepancies. From the results, we observe that our framework successfully identifies constraint violations and generates appropriate corrections that restore spatial coherence and design consistency. The corrected layouts maintain natural spatial relationships while satisfying the specified constraints, demonstrating the method's effectiveness in addressing various constraint types simultaneously. These results show that our approach can successfully resolve common constraint violations encountered in indoor scene generation.

\subsubsection{Quantitative Results.}

\noindent\textbf{Layout quality results.} To evaluate our method against 3D indoor room generation models, we conducted experiments to compare with current state-of-the-art methods, including InsctrucScene \cite{InstructScene}, Holodeck \cite{Holodeck}, LayoutGPT \cite{LayoutGPT}, AnyHome \cite{AnyHome}, Infinigen Indoor \cite{InfinigenIndoor} and Chat2Layout \cite{Chat2Layout}. To compare quality of synthesized scenes, we report the standard metrics like OOB (out-of-bound rate), ORI (orientation correctness), and CLIP-Sim (CLIP similarity) following Chat2Layout \cite{Chat2Layout}. 
As shown in Table \ref{tab:OOB}, our approach achieved the best results across all metrics, demonstrating significant improvements in spatial coherence, functionality, and alignment with text descriptions. With the lowest OOB of 19.3, our model ensures spatial efficiency and accurate object placement within scene boundaries. The ORI score of 89.3 outperforms all other methods, indicating better alignment with room layout conventions. The CLIP-Sim of 27.7 reflects semantic alignment with input descriptions. 

\begin{figure}[!tbp]
    \centering
    \includegraphics[width=0.98\linewidth]{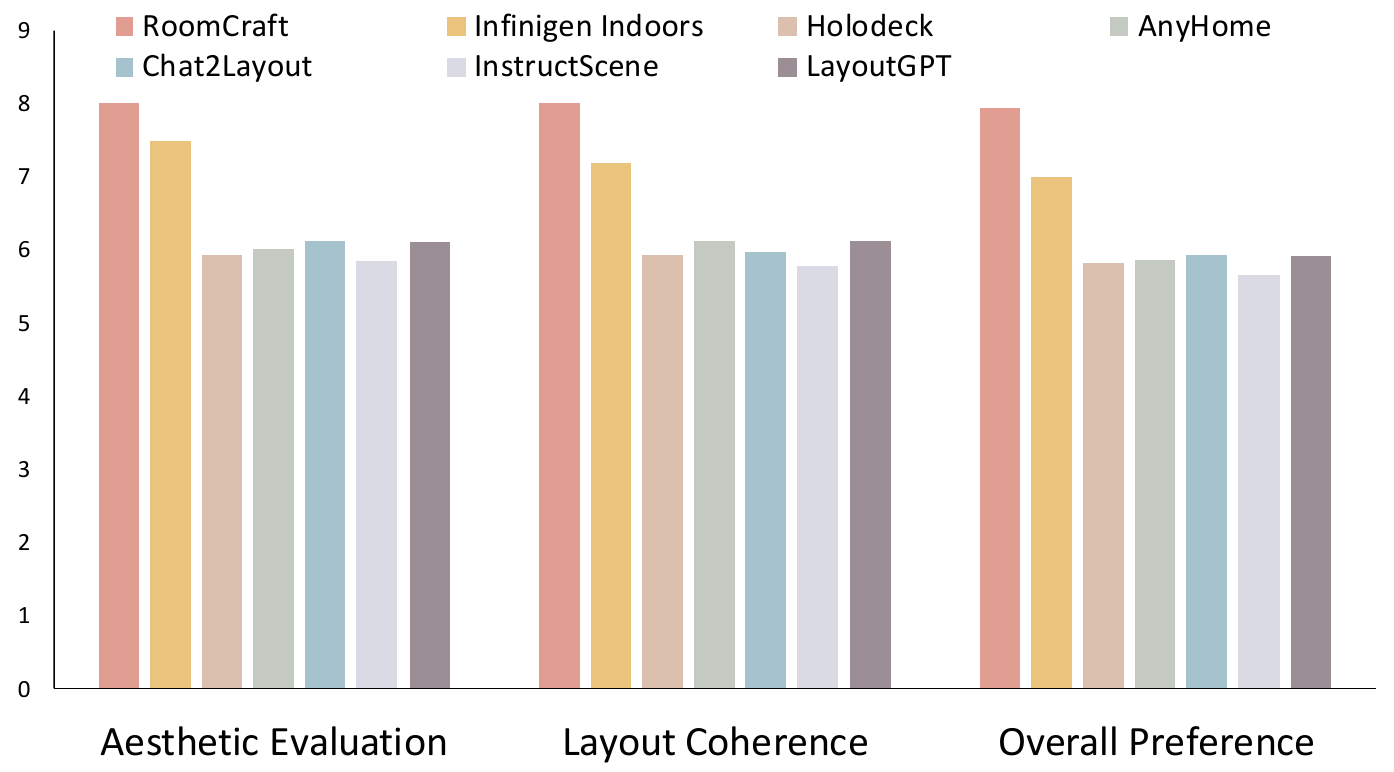}
    \caption{Human evaluation results for scene quality in RoomCraft, including Aesthetic Evaluation, Layout Coherence, and Overall Preference.}
    \label{fig:userstudy}
\end{figure}

\noindent\textbf{User Study.}
    To assess the quality of generated scenes from RoomCraft, we conduct human evaluations in this subsection. We provide three subjective evaluation metrics, including \textit{Aesthetic Evaluation}, \textit{Layout Coherence}, and \textit{Overall Preference}. Aesthetic evaluation assesses the quality of the generated scenes from the visual appeal, harmony, and overall aesthetics of the scenes on a standardized scale. Layout coherence refers to the logical consistency of the spatial relationships and orientations of objects, as well as the alignment between the generated scene and the given prompt. Moreover, \textit{Overall Preference} reveals which generated scene the participants prefer among the compared methods. We count the scores from 100 graduate students participating and the results are shown in Figure \ref{fig:userstudy}. From the results, we can observe that RoomCraft excels across all three evaluation metrics. In Aesthetic Evaluation, RoomCraft scores the highest at 8.026, significantly outperforming the other methods. For Layout Coherence, RoomCraft also leads with a score of 8.008, indicating more logical spatial relationships and object orientations in the generated scenes. In Overall Preference, RoomCraft receives the highest preference score of 7.96, showing participants' inclination towards scenes generated by RoomCraft.

\begin{figure}[!tbp]
    \centering
    \includegraphics[width=1\linewidth]{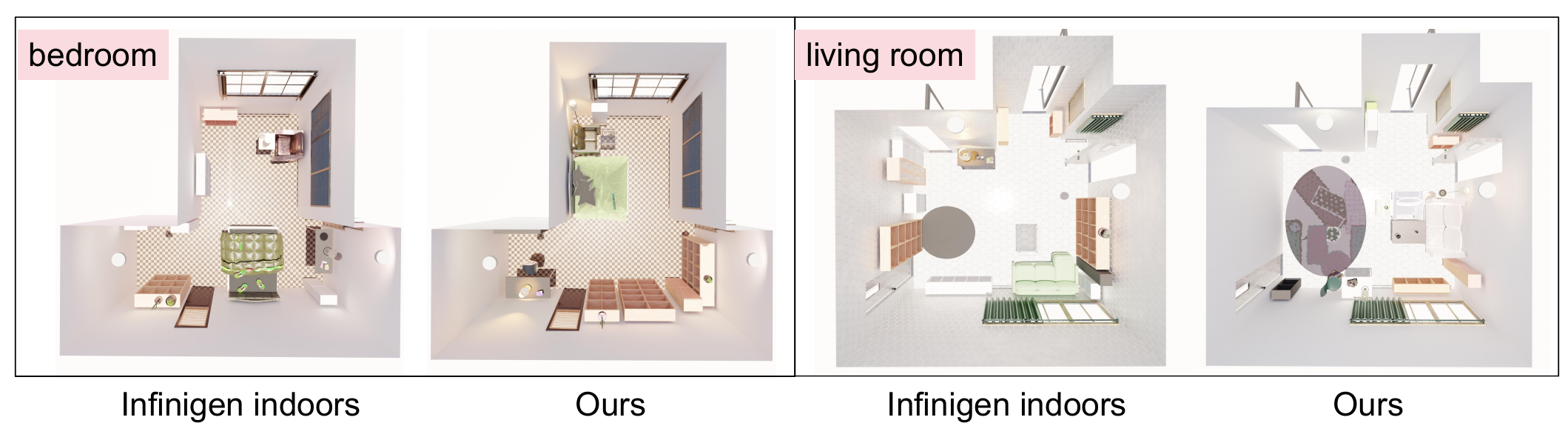}
    \caption{Qualitative Comparison of generated indoor scenes between RoomCraft and Infinigen Indoors. 
    }
    \label{fig:ablation_vs_infinigen}
    \vspace{-0.5 cm}
\end{figure}

\subsection{Ablation Study}
This ablation study investigates the contributions of two key modules in RoomCraft’s layout generation: the Heuristic-based Depth-First Search (HDFS) and Conflict-Aware Positioning Strategy (CAPS).

\noindent\textbf{The influence of CAPS.} To evaluate the conflict-aware positioning strategy, we use quantitative evaluations using metrics such as OOB (out-of-bound rate), ORI (orientation correctness), and CLIP-Sim (CLIP similarity). The corresponding results are shown in Table \ref{tab:OOB}. Compared to the model without CAPS, the full model achieves a significant improvement, specifically for OOB and ORI metrics.

\begin{figure}[!tbp]
    \centering
    \includegraphics[width=1\linewidth]{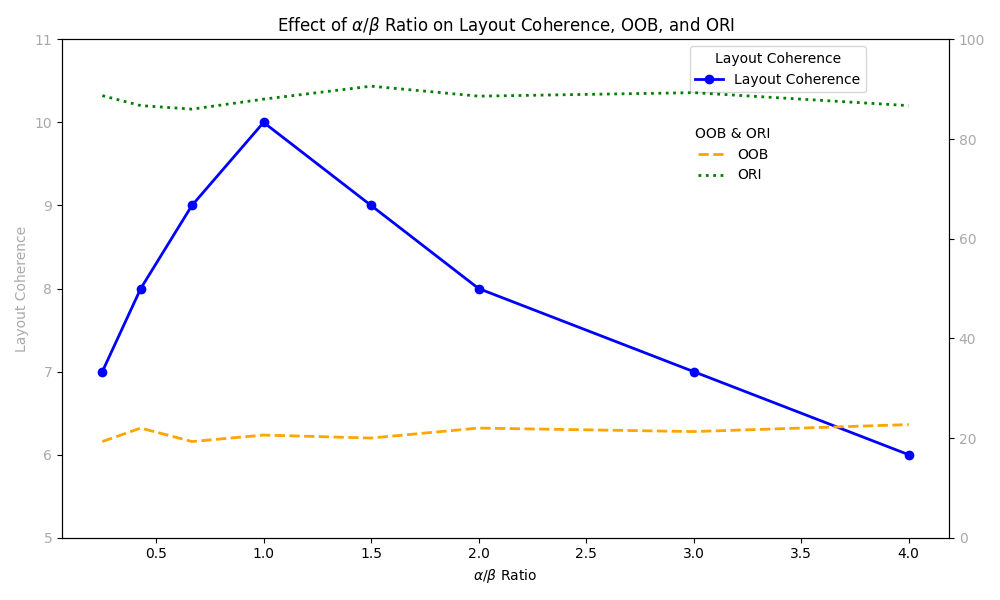}
    \caption{Parameters analysis for the ratio of $\alpha$ and $\beta$.}
    \label{fig:param}
    \vspace{-1 cm}
\end{figure}

\noindent\textbf{The influence of HDFS. }
To assess RoomCraft’s effectiveness, we provide prompts to guide the generation of indoor scenes. For a fair comparison with Infinigen Indoor \cite{InfinigenIndoor}, we manually craft corresponding input scripts, ensuring consistent information across both systems. The results are shown in Figure \ref{fig:ablation_vs_infinigen}. As observed, Infinigen's generated indoor scenes are less rich than ours. This is due to Infinigen's inability to fully adhere to the input prompts, which often results in missing key elements. In contrast, our framework ensures the inclusion of all specified furniture and objects, leading to more complete and consistent scene generation.

\noindent\textbf{Parameter Analysis.} In this experiment, we examine the impact of the ratio between the hyper-parameters $\alpha$ and $\beta$ on three metrics: Layout Coherence, Out of Bounds (OOB), and Object Rotation Index (ORI). The results, summarized in Figure \ref{fig:param}, show that the ratio of $\alpha$ to $\beta$ significantly affects Layout Coherence, with the highest score achieved when the ratio is around 1.0. This indicates that a balanced $\alpha$ to $\beta$ ratio is crucial for maintaining spatial consistency and alignment with the prompt. However, the ratio has minimal effect on OOB and ORI.

%% file: sec/5_Conclusions.tex
\section{Conclusions}
In this paper, we introduced RoomCraft, a novel system for generating realistic 3D indoor scenes from diverse user inputs including real images, sketches, or text descriptions. By leveraging VLM and a multi-stage pipeline, RoomCraft effectively extracts and structures scene information to ensure input consistency. The constraint-driven optimization framework systematically handles spatial, functional, and aesthetic requirements, while our conflict-aware positioning strategy (CAPS) significantly reduces furniture placement conflicts through dynamic adjustment.

\noindent\textbf{Limitations. }
RoomCraft has two primary limitations. First, its performance is inherently constrained by the capabilities of pre-trained VLM, which may lead to suboptimal results when handling novel or highly complex indoor layouts. Second, the accuracy of furniture placement is dependent on the granularity of predefined spatial relationships, limiting the system's ability to achieve fine-grained control over object positioning.

%% file: sec/6_suppl.tex
\clearpage
\setcounter{page}{1}

\section*{A: Comparison with Infinigen Indoors}

Infinigen represents a significant advancement in Procedural Content Generation (PCG) for indoor scenes, offering a comprehensive asset library and basic furniture layout capabilities, our analysis reveals several critical limitations in its practical application. 
The key limitations include:
\begin{itemize}
    \item \textbf{Inconsistent Asset Preservation:} The system frequently fails to maintain the specified number of furniture items due to unresolved spatial conflicts during generation.
    \item \textbf{Limited Spatial Reasoning:} The algorithm can only handle basic positioning patterns with 0-2 degrees of freedom, struggling with complex spatial relationships.
    \item \textbf{Poor Scene Coherence:} Generated environments often exhibit physical interference between objects or lack logical spatial arrangements.
    \item \textbf{Inflexible Constraint Handling:} The system struggles to accommodate scenarios requiring precise spatial configurations or multiple inter-object constraints.
\end{itemize}

\begin{figure*}[!tbp]
    \centering
    \includegraphics[width=0.98\linewidth]{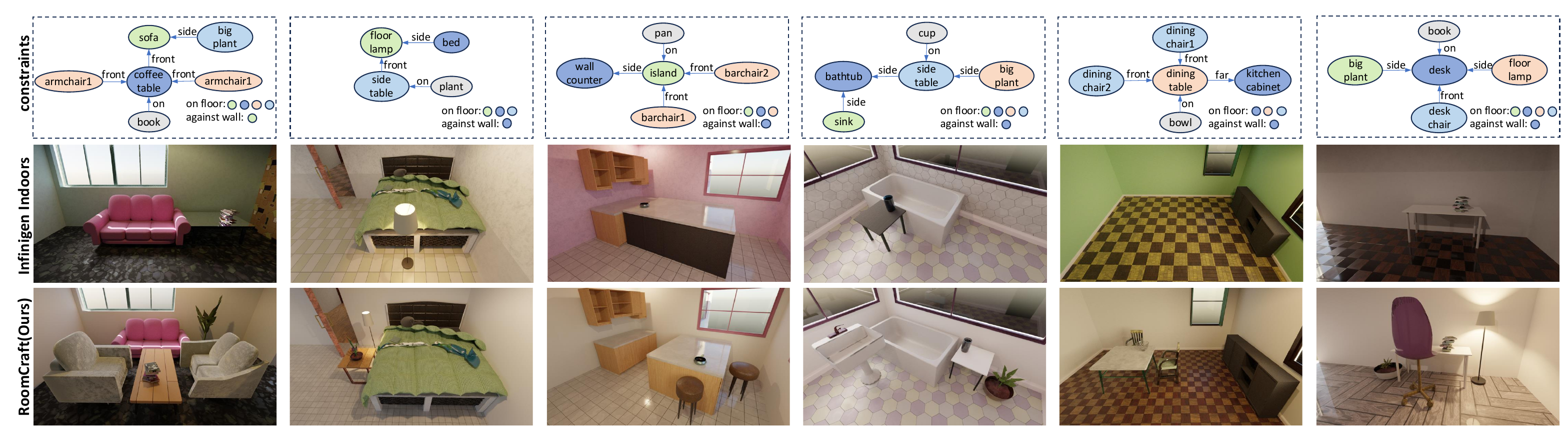}
    \caption{Visual comparison between Infinigen and RoomCraft.}
    \label{fig:visual comparison}
\end{figure*}

\noindent To address these limitations, RoomCraft introduces several innovative solutions:
\begin{itemize}
    \item \textbf{To address limited spatial reasoning and inflexible constraints:} We propose a heuristic depth-first search algorithm that decomposes complex indoor layouts into independent furniture placement subtasks. This decomposition strategy, guided by our carefully designed heuristic function, determines optimal placement priorities while handling spatial constraints with high degrees of freedom, effectively overcoming Infinigen's limitations in handling complex spatial relationships.
    
    \item \textbf{To solve asset preservation and scene coherence issues:} We introduce an adaptive layout optimization mechanism that establishes comprehensive evaluation criteria for placement schemes. Through dynamic weight adjustment and iterative optimization, our system maintains the desired number of furniture items while ensuring their spatial relationships remain both natural and practical. The evaluation considers multiple dimensions including spatial utilization, aesthetic appeal, and functionality, directly addressing the physical interference and logical arrangement problems present in Infinigen.
\end{itemize}
To demonstrate these improvements, we present a comparative analysis between RoomCraft and Infinigen in Fig \ref{fig:visual comparison}. With six distinct examples, the figure shows input constraints (first row), Infinigen's results (second row), and RoomCraft's outputs (third row). Infinigen consistently exhibits limitations: it fails to preserve complete furniture sets (e.g., missing armchairs, coffee table, and books in the first example) and struggles with spatial relationships (e.g., incorrectly positioning a floor lamp in front of rather than beside the bed in the second example). In contrast, RoomCraft successfully preserves all furniture items while maintaining accurate spatial relationships across all test cases, validating the effectiveness of our proposed solutions.



\setcounter{subsection}{0} 
\renewcommand{\thesubsection}{B.\arabic{subsection}}
\section*{B: Technical Details}
\subsection{Furniture Relationship}
In the domain of indoor scene generation, precise spatial relationship modeling is crucial for creating functional and realistic environments. Our system captures furniture-space interactions through a multi-dimensional relationship framework:

\begin{itemize}
\item Room Type
    \begin{itemize}
    \item Living Room: Social and entertainment space
    \item Bathroom: Personal hygiene area
    \item Dining Room: Eating space
    \item Kitchen: Food preparation area
    \item Bedroom: Resting space
    \end{itemize}

\item Room Relationship
    \begin{itemize}
    \item Floor Placement: On-floor objects
    \item Wall Relationship: Adjacent or distant to walls
    \item Ceiling Mount: Suspended objects
    \item Combined Constraints: Floor-based with wall specifications
    \end{itemize}

\item Inter-Furniture Relationships
    \begin{itemize}
    \item Directional: in front of, left/right to
    \item Orientational: face to face, back to back
    \item Positional: side by side, aligned with
    \item Vertical: on top of
    \end{itemize}

\item Distance Constraints
    \begin{itemize}
    \item Contact: Direct physical contact
    \item Proximity: Near placement
    \item Distance: Separated placement
    \item Numerical: Specific measurements
    \end{itemize}
\end{itemize}

These relationships form our semantic framework for generating coherent and reasonable indoor scenes.

\begin{algorithm}[!h]
    \caption{The workflow of RoomCraft}
    \label{alg:roomcraft}
    \LinesNumbered
    \KwIn{Input source $I$ (sketches/images/text), MLLM model $M$, Prompts $P_I$, Scene organization $O$, furniture collection $V$, spatial relations $E$, $End\_flag = False$}
    \KwOut{Final 3D room scene}
    When receiving an input source, the RoomCraft system starts the scene generation workflow\;
    
    \While{$End\_flag = False$}{
        Process input using MLLM to extract scene information\;
        $O = P_{GPT-4o}(I, P_I)$\;
        
        \For{each component in $O$}{
            $\{R, V, E\} = \text{ParseSceneOrganization}(O)$\;
            Analyze room type, furniture items and spatial relations\;
            
            \For{each pair $(V_i, V_j)$ in $V$}{
                Constraint $\pi_{i,j} = \text{CreateConstraint}(V_i, V_j, E_{i,j})$\;
                Build spatial relationship graph\;
                
                \eIf{spatial relation exists}{
                    Calculate heuristic cost $f(V_i) = \sum_{j=1}^{m} w_i \cdot \mathbb{I}_j(V_i, V_j, E_{i,j})$\;
                }{
                    Continue to next pair\;
                }
                
                \If{All constraints are processed}{
                    Sort furniture by heuristic cost\;
                    Generate ordered layout $\{V_1', V_2', ..., V_m'\}$\;
                    Generate basic room structure\;
                    Add furniture according to ordered layout\;
                    $End\_flag \leftarrow True$\;
                }
            }
        }
    }
    Ends when the complete 3D room scene is generated\;
\end{algorithm}

\subsection{Implementation Details and Examples}
\begin{figure*}[!tbp]
    \centering
    \includegraphics[width=0.98\linewidth]{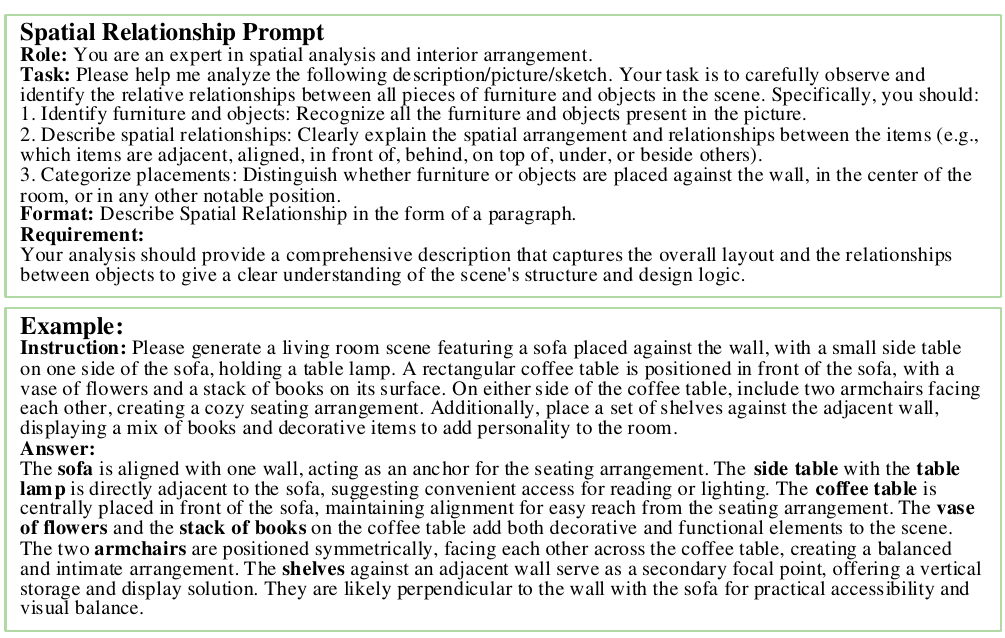}
    \caption{Prompt example for spatial relationship analysis in indoor scenes.}
    \label{fig:spatial relationship}
\end{figure*}

\begin{figure*}[!tbp]
    \centering
    \includegraphics[width=0.8\linewidth]{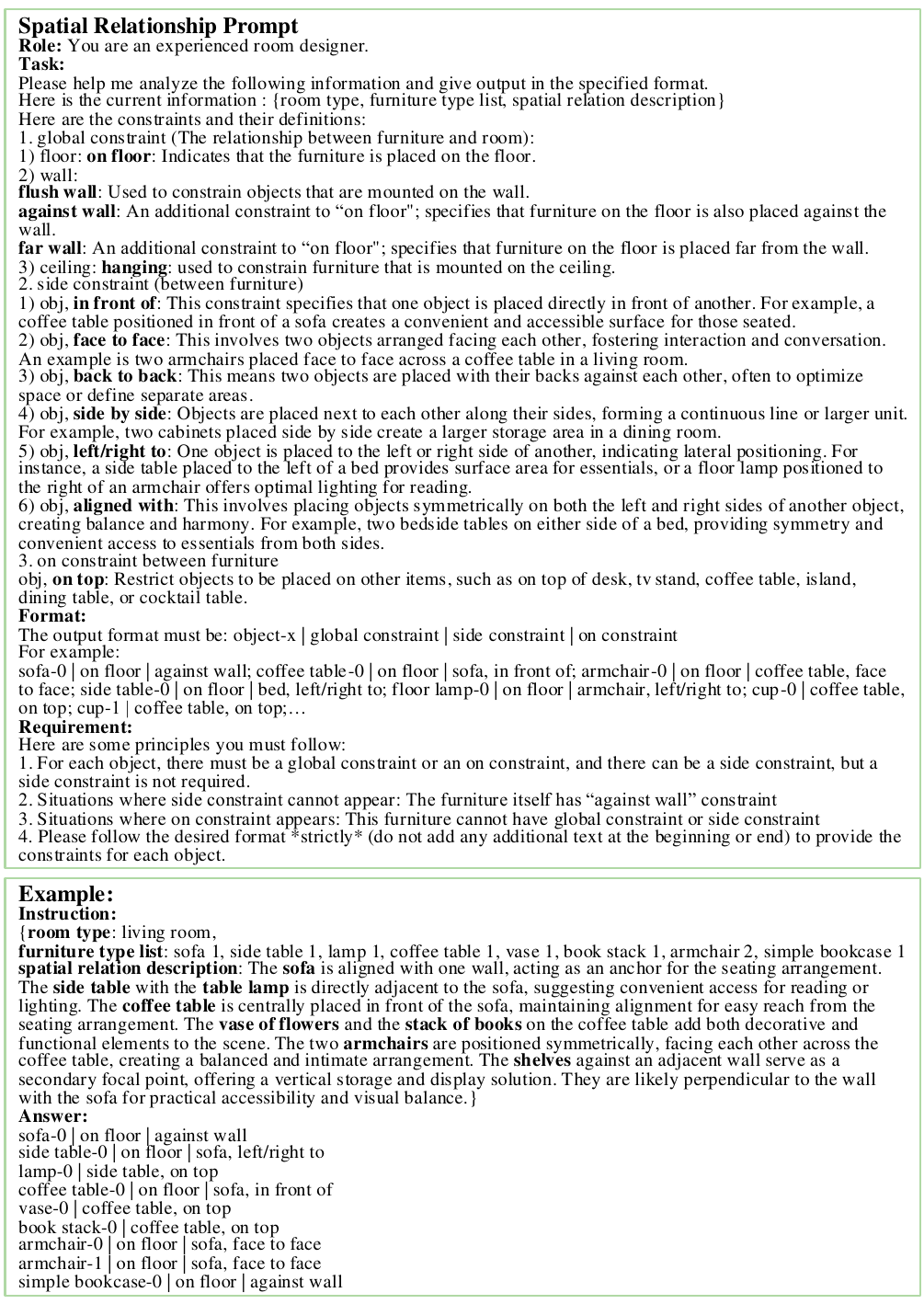}
    \caption{Prompt example for constraint-based furniture relationship formalization.}
    \label{fig:furniture relationship}
\end{figure*}

\begin{figure*}[!tbp]
    \centering
    \includegraphics[width=0.8\linewidth]{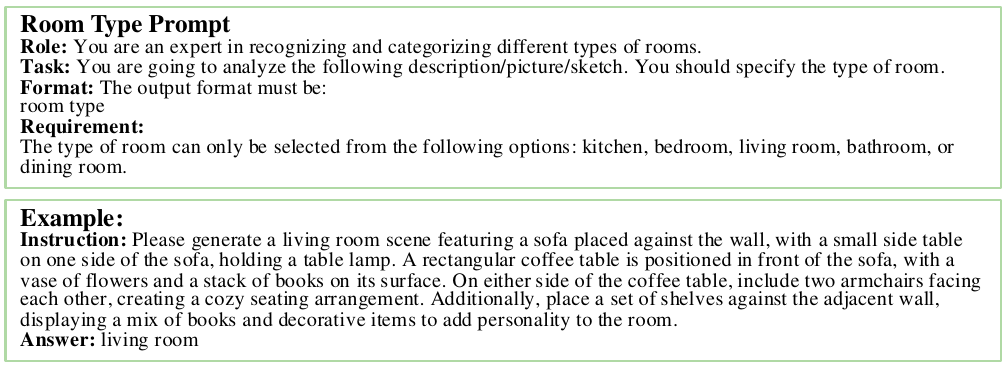}
    \vspace{-0.2cm}
    \caption{Prompt example for room type classification.}
    \label{fig:room type}
\end{figure*}

\begin{figure*}[!tbp]
    \centering
    \includegraphics[width=0.8\linewidth]{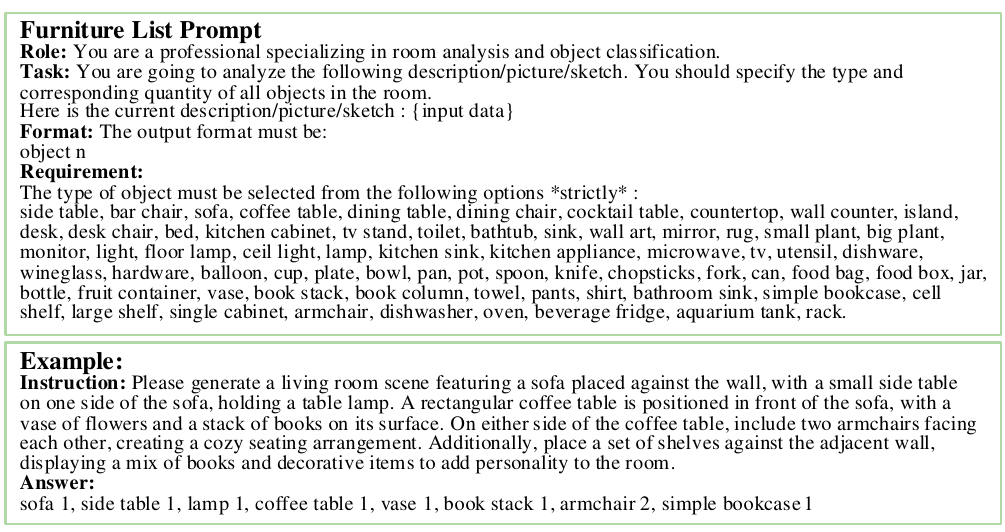}
    \caption{Prompt example for furniture enumeration and quantification.}
    \label{fig:furniture enumeration}
    \vspace{-0.5cm}
\end{figure*}

To provide a comprehensive understanding of our implementation process, we present a series of detailed examples and code demonstrations. Figures~\ref{fig:spatial relationship}-\ref{fig:furniture enumeration} illustrate various stages of our prompt design and execution, showcasing different prompt templates and their corresponding examples for room type specification, furniture enumeration, spatial relationship analysis, and constraint-based formalization.

\begin{figure*}[!tbp]
    \centering
    \includegraphics[width=0.8\linewidth]{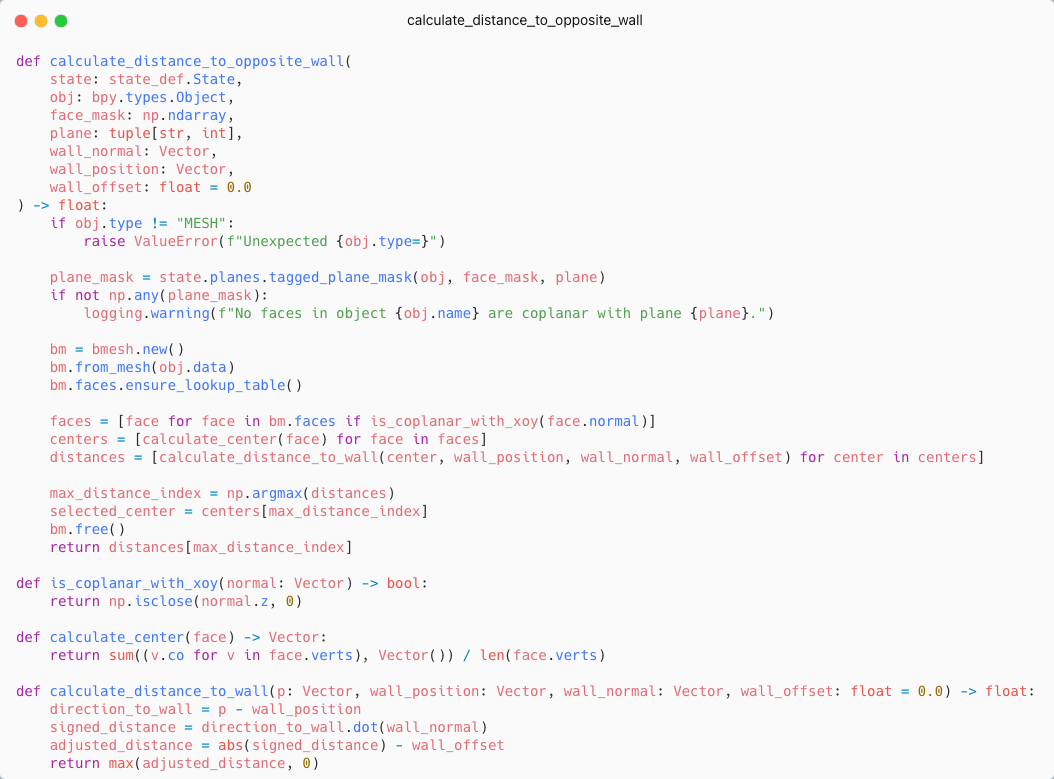}
    \vspace{-0.2cm}
    \caption{Implementation of the wall distance constraint function.}
    \label{fig:code1}
\end{figure*}

\begin{figure*}[!tbp]
    \centering
    \includegraphics[width=0.8\linewidth]{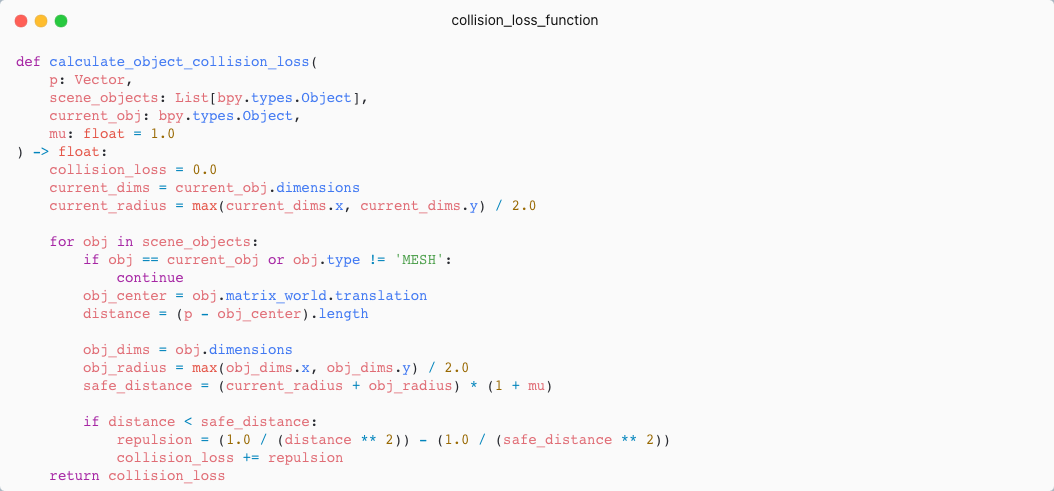}
    \vspace{-0.2cm}
    \caption{Implementation of the object density component.}
    \label{fig:code3}
    \vspace{-0.5cm}
\end{figure*}

The implementation of our furniture placement optimization is demonstrated through the code examples in Figures~\ref{fig:code1} and~\ref{fig:code3}. These figures showcase the computational details of two key components in our objective function: $\mathcal{L}_{\text{dist}}(p)$ and $\mathcal{L}_{\text{obj}}(p,\mu)$. The $\mathcal{L}_{\text{dist}}(p)$ component, illustrated in Figure~\ref{fig:code1}, computes the distance-based constraints to ensure appropriate spacing between furniture and walls. Meanwhile, Figure~\ref{fig:code3} presents the implementation of $\mathcal{L}_{\text{obj}}(p,\mu)$, which calculates object-to-object relationships to minimize interference between furniture pieces. Together, these components form the foundation of our placement optimization algorithm, enabling precise and practical furniture arrangements.

To further illustrate our system's workflow, we present Algorithm \ref{alg:roomcraft}, which outlines the complete process of RoomCraft from input processing to final scene generation.

\setcounter{subsection}{0} 
\renewcommand{\thesubsection}{B.\arabic{subsection}}
\section*{C: Visualization Results}
\begin{figure*}[!tbp]
    \centering
    \includegraphics[width=0.98\linewidth]{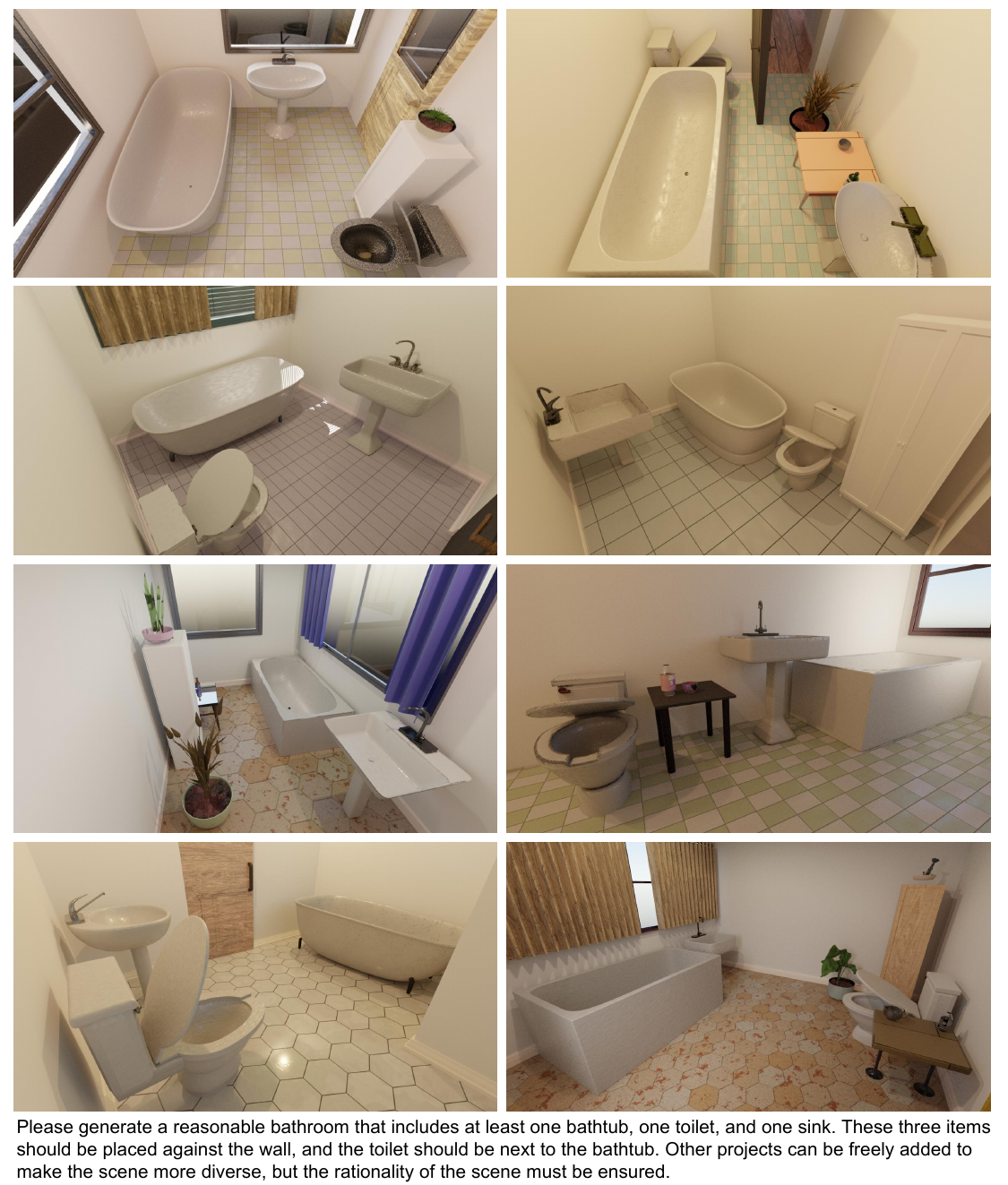}
    \vspace{-0.2cm}
    \caption{Diverse 3D indoor scene generation through varied input text.}
    \label{fig:demo1}
    \vspace{-0.5cm}
\end{figure*}

\begin{figure*}[!tbp]
    \centering
    \includegraphics[width=0.98\linewidth]{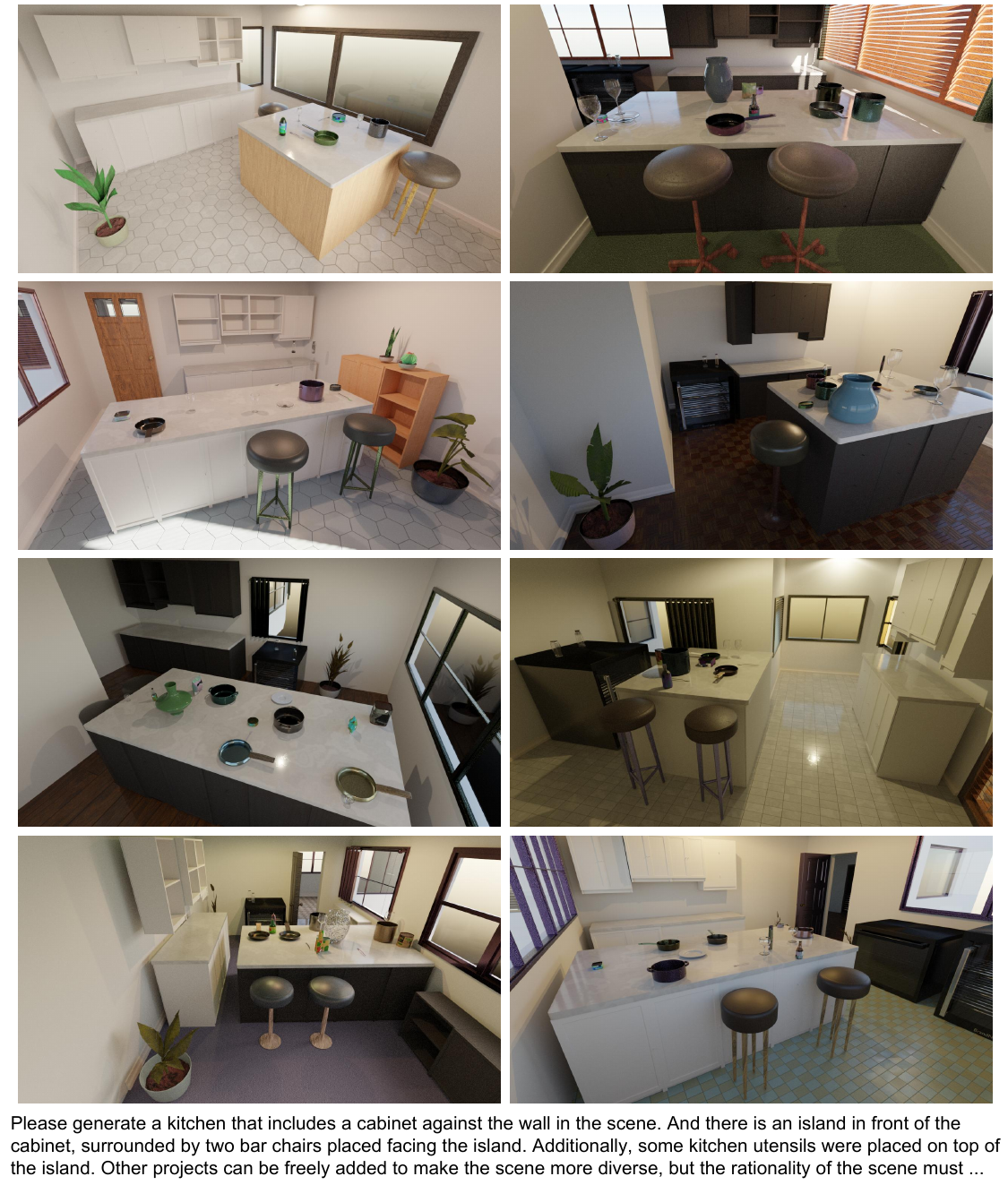}
    \vspace{-0.2cm}
    \caption{Diverse 3D indoor scene generation through varied input text.}
    \label{fig:demo2}
    \vspace{-0.5cm}
\end{figure*}

\begin{figure*}[!tbp]
    \centering
    \includegraphics[width=0.98\linewidth]{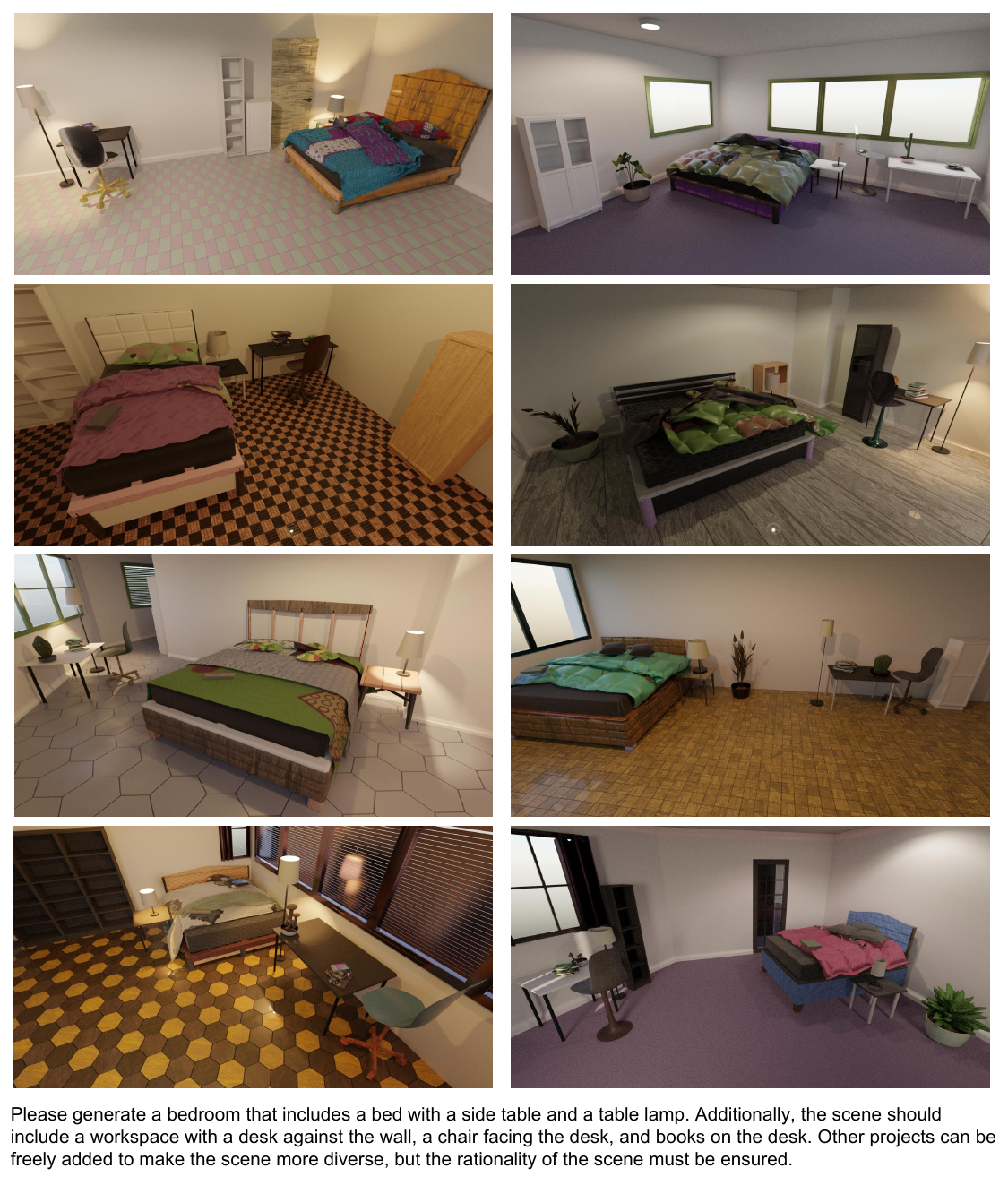}
    \vspace{-0.2cm}
    \caption{Diverse 3D indoor scene generation through varied input text.}
    \label{fig:demo3}
    \vspace{-0.5cm}
\end{figure*}

\begin{figure*}[!tbp]
    \centering
    \includegraphics[width=0.98\linewidth]{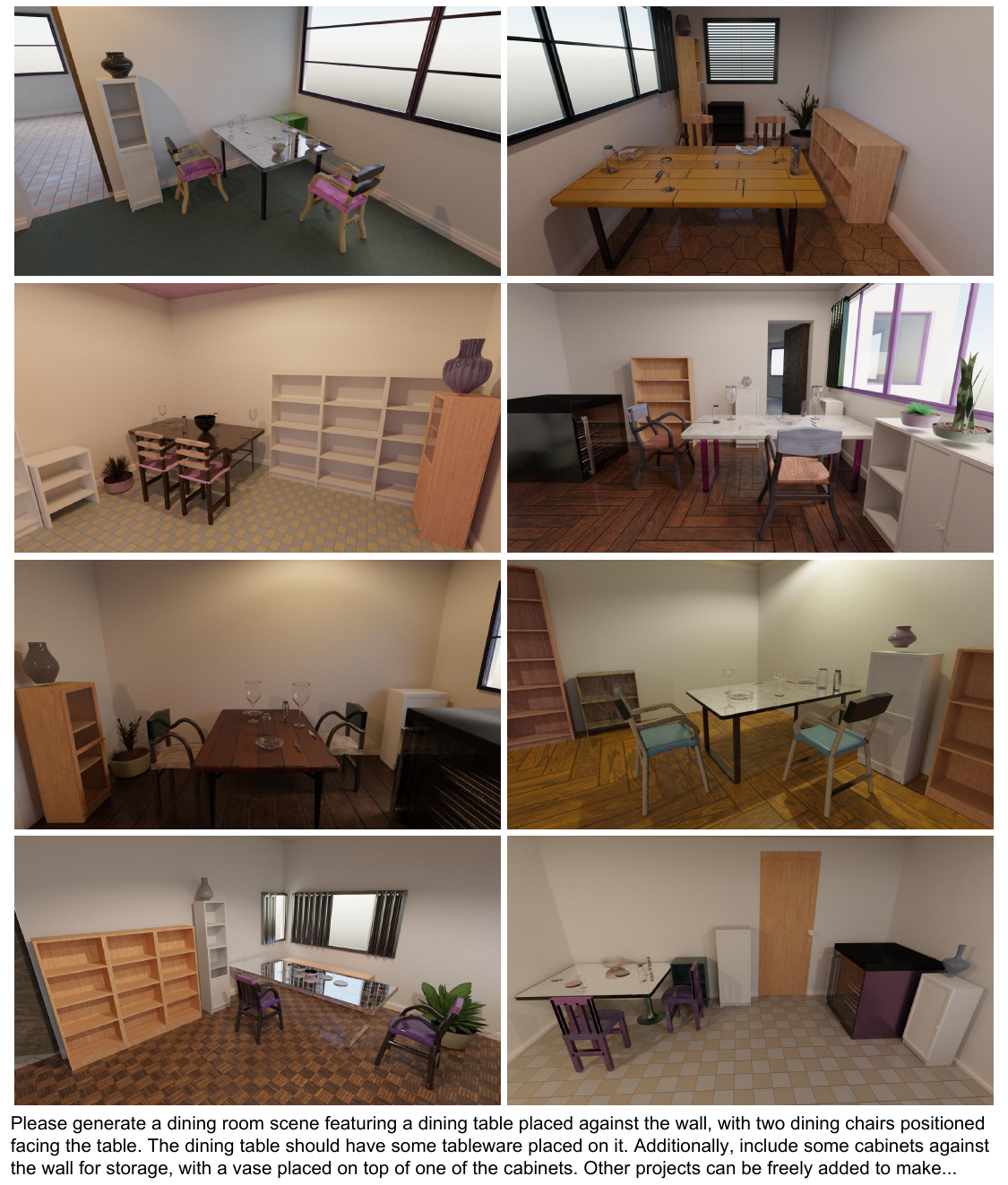}
    \vspace{-0.2cm}
    \caption{Diverse 3D indoor scene generation through varied input text.}
    \label{fig:demo4}
    \vspace{-0.5cm}
\end{figure*}

\begin{figure*}[!tbp]
    \centering
    \includegraphics[width=0.98\linewidth]{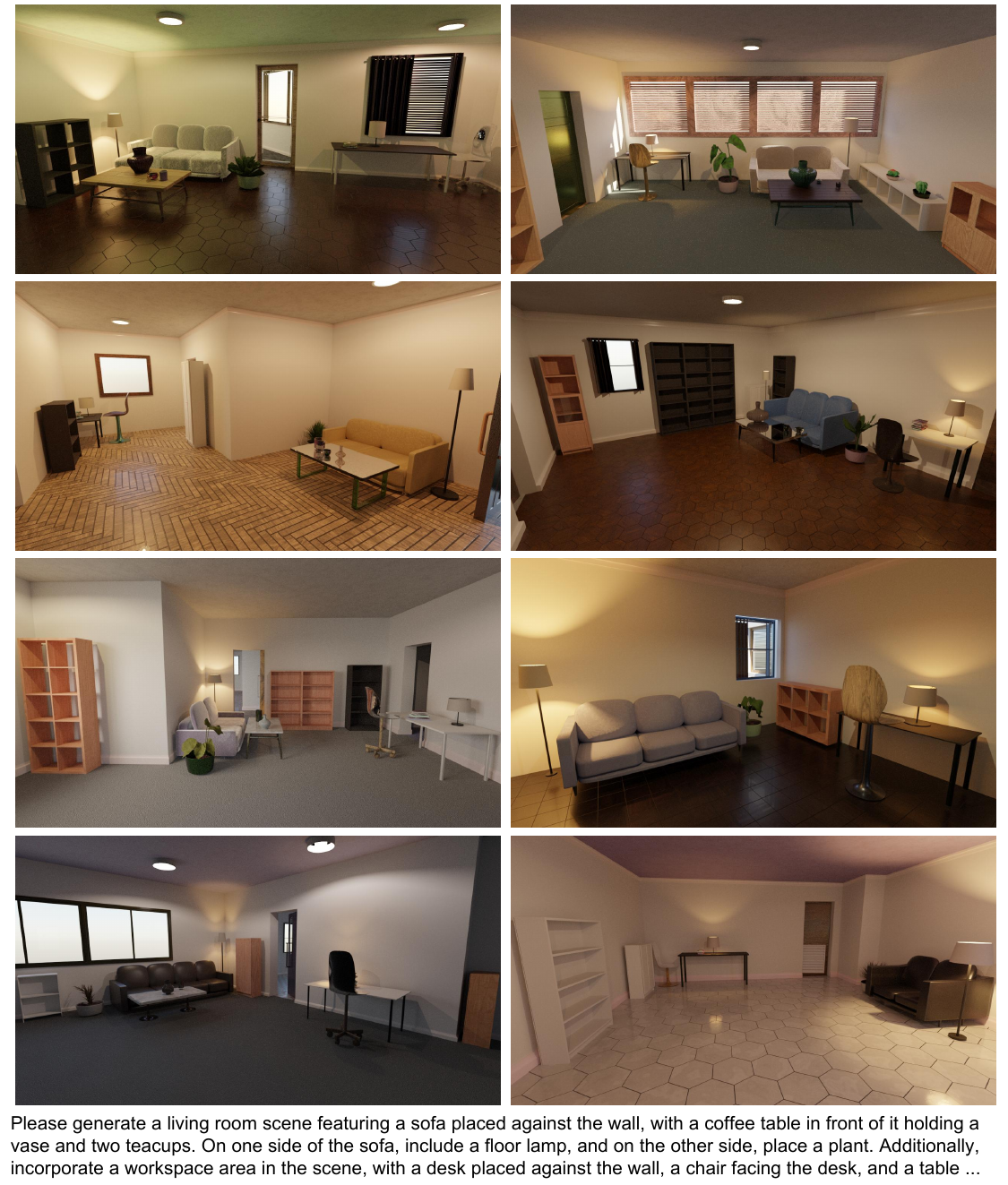}
    \vspace{-0.2cm}
    \caption{Diverse 3D indoor scene generation through varied input text.}
    \label{fig:demo5}
    \vspace{-0.5cm}
\end{figure*}

\begin{figure*}[!tbp]
    \centering
    \includegraphics[width=0.98\linewidth]{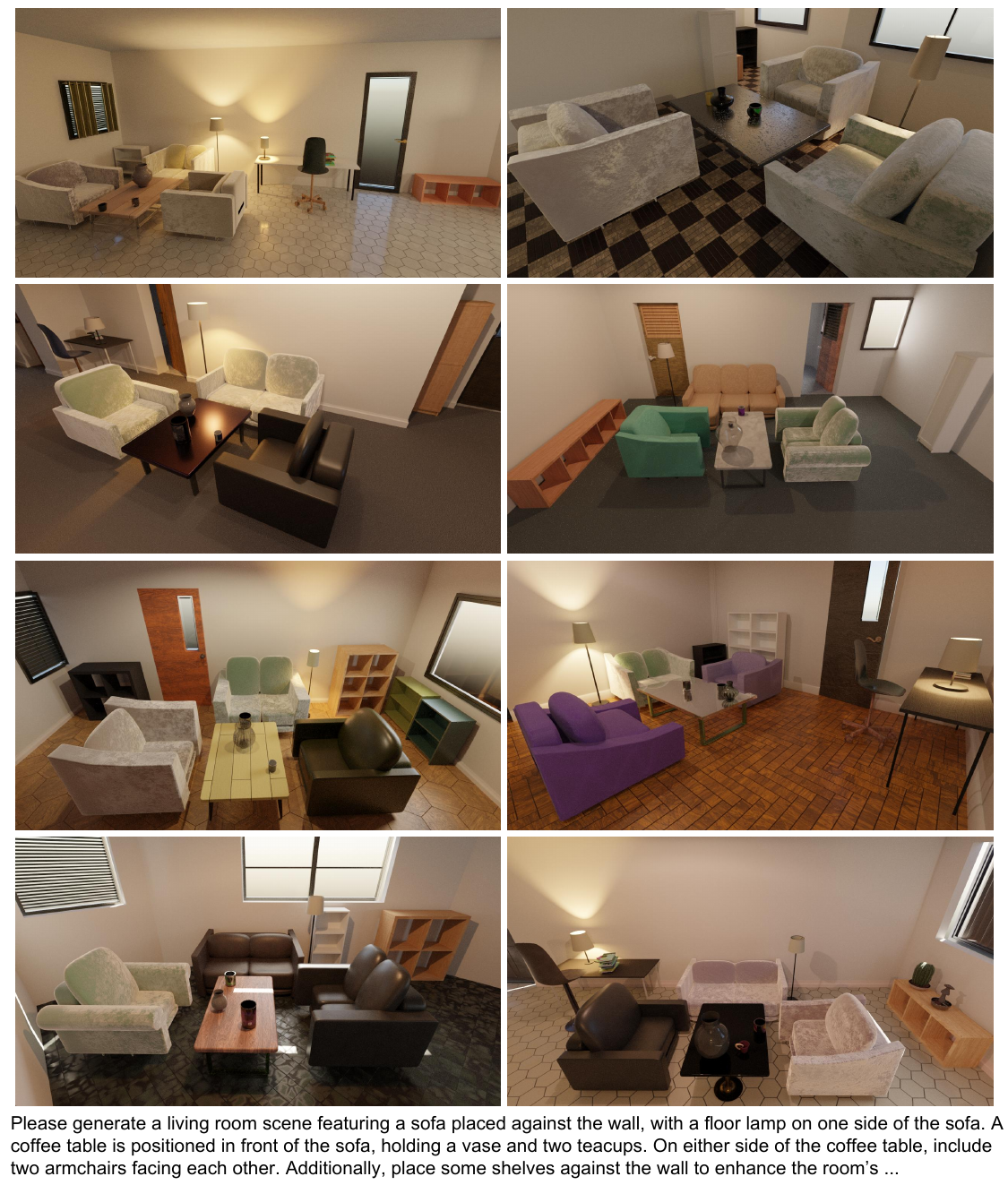}
    \vspace{-0.2cm}
    \caption{Diverse 3D indoor scene generation through varied input text.}
    \label{fig:demo6}
    \vspace{-0.5cm}
\end{figure*}

\begin{figure*}[!tbp]
    \centering
    \includegraphics[width=0.98\linewidth]{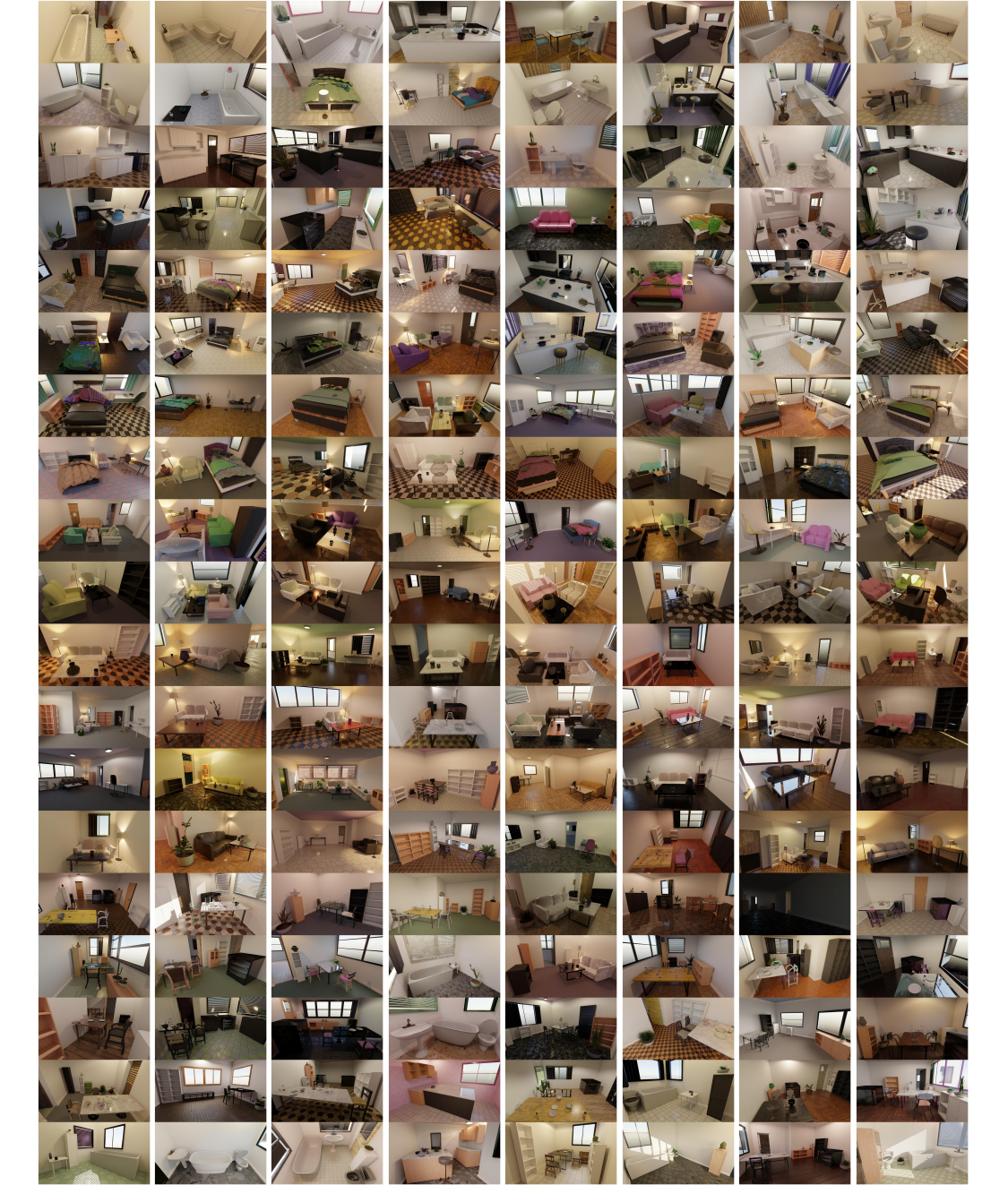}
    \vspace{-0.2cm}
    \caption{Gallery of generated 3D indoor scenes.}
    \label{fig:demo7}
    \vspace{-0.5cm}
\end{figure*}

We present our visualization results through multiple examples. Figures~\ref{fig:demo1}-\ref{fig:demo6} demonstrate our system's ability to generate 3D indoor scenes from different text inputs. Additionally, Figure~\ref{fig:demo7} shows a gallery of diverse generated scenes, further validating the effectiveness of our approach.

%% file: main.bbl

\begin{thebibliography}{46}


\ifx \showCODEN    \undefined \def \showCODEN     #1{\unskip}     \fi
\ifx \showDOI      \undefined \def \showDOI       #1{#1}\fi
\ifx \showISBNx    \undefined \def \showISBNx     #1{\unskip}     \fi
\ifx \showISBNxiii \undefined \def \showISBNxiii  #1{\unskip}     \fi
\ifx \showISSN     \undefined \def \showISSN      #1{\unskip}     \fi
\ifx \showLCCN     \undefined \def \showLCCN      #1{\unskip}     \fi
\ifx \shownote     \undefined \def \shownote      #1{#1}          \fi
\ifx \showarticletitle \undefined \def \showarticletitle #1{#1}   \fi
\ifx \showURL      \undefined \def \showURL       {\relax}        \fi
\providecommand\bibfield[2]{#2}
\providecommand\bibinfo[2]{#2}
\providecommand\natexlab[1]{#1}
\providecommand\showeprint[2][]{arXiv:#2}

\bibitem[Cabral and Furukawa(2014)]%
        {Cabral14}
\bibfield{author}{\bibinfo{person}{R. Cabral} {and} \bibinfo{person}{Y. Furukawa}.} \bibinfo{year}{2014}\natexlab{}.
\newblock \showarticletitle{Piecewise planar and compact floorplan reconstruction from images}. In \bibinfo{booktitle}{\emph{2014 IEEE Conference on Computer Vision and Pattern Recognition}}. IEEE.
\newblock


\bibitem[Chen et~al\mbox{.}(2023a)]%
        {Fantasia3d}
\bibfield{author}{\bibinfo{person}{Rui Chen}, \bibinfo{person}{Yongwei Chen}, \bibinfo{person}{Ningxin Jiao}, {and} \bibinfo{person}{Kui Jia}.} \bibinfo{year}{2023}\natexlab{a}.
\newblock \showarticletitle{Fantasia3d: Disentangling geometry and appearance for high-quality text-to-3d content creation}.
\newblock \bibinfo{journal}{\emph{arXiv preprint arXiv:2303.13873}} (\bibinfo{year}{2023}).
\newblock


\bibitem[Chen et~al\mbox{.}(2023b)]%
        {SceneDreamer}
\bibfield{author}{\bibinfo{person}{Zhaoxi Chen}, \bibinfo{person}{Guangcong Wang}, {and} \bibinfo{person}{Ziwei Liu}.} \bibinfo{year}{2023}\natexlab{b}.
\newblock \showarticletitle{Scenedreamer: Unbounded 3d scene generation from 2d image collections}.
\newblock \bibinfo{journal}{\emph{arXiv preprint arXiv:2302.01330}} (\bibinfo{year}{2023}).
\newblock


\bibitem[Dai et~al\mbox{.}(2024a)]%
        {ACDC}
\bibfield{author}{\bibinfo{person}{Tianyuan Dai}, \bibinfo{person}{Josiah Wong}, \bibinfo{person}{Yunfan Jiang}, \bibinfo{person}{Chen Wang}, \bibinfo{person}{Cem Gokmen}, \bibinfo{person}{Ruohan Zhang}, \bibinfo{person}{Jiajun Wu}, {and} \bibinfo{person}{Li Fei-Fei}.} \bibinfo{year}{2024}\natexlab{a}.
\newblock \showarticletitle{Acdc: Automated creation of digital cousins for robust policy learning}.
\newblock \bibinfo{journal}{\emph{arXiv preprint arXiv:2410.07408}} (\bibinfo{year}{2024}).
\newblock


\bibitem[Dai et~al\mbox{.}(2024b)]%
        {Dai24}
\bibfield{author}{\bibinfo{person}{Tianyuan Dai}, \bibinfo{person}{Josiah Wong}, \bibinfo{person}{Yunfan Jiang}, \bibinfo{person}{Chen Wang}, \bibinfo{person}{Cem Gokmen}, \bibinfo{person}{Ruohan Zhang}, \bibinfo{person}{Jiajun Wu}, {and} \bibinfo{person}{Li Fei-Fei}.} \bibinfo{year}{2024}\natexlab{b}.
\newblock \showarticletitle{Automated Creation of Digital Cousins for Robust Policy Learning}.
\newblock \bibinfo{journal}{\emph{arXiv preprint arXiv:2410.07408}} (\bibinfo{year}{2024}).
\newblock
\urldef\tempurl%
\url{https://doi.org/10.48550/arXiv.2410.07408}
\showURL{%
\tempurl}
\newblock
\shownote{CoRL 2024}.


\bibitem[Deitke et~al\mbox{.}(2022)]%
        {ProcTHOR}
\bibfield{author}{\bibinfo{person}{Matt Deitke}, \bibinfo{person}{Eli VanderBilt}, \bibinfo{person}{Alvaro Herrasti}, \bibinfo{person}{Luca Weihs}, \bibinfo{person}{Kiana Ehsani}, \bibinfo{person}{Jordi Salvador}, \bibinfo{person}{Winson Han}, \bibinfo{person}{Eric Kolve}, \bibinfo{person}{Aniruddha Kembhavi}, {and} \bibinfo{person}{Roozbeh Mottaghi}.} \bibinfo{year}{2022}\natexlab{}.
\newblock \showarticletitle{ProcTHOR: Large-Scale Embodied AI Using Procedural Generation}.
\newblock \bibinfo{journal}{\emph{Advances in Neural Information Processing Systems}}  \bibinfo{volume}{35} (\bibinfo{year}{2022}), \bibinfo{pages}{5982--5994}.
\newblock


\bibitem[Feng et~al\mbox{.}(2024)]%
        {LayoutGPT}
\bibfield{author}{\bibinfo{person}{Weixi Feng}, \bibinfo{person}{Wanrong Zhu}, \bibinfo{person}{Tsu-jui Fu}, \bibinfo{person}{Varun Jampani}, \bibinfo{person}{Arjun Akula}, \bibinfo{person}{Xuehai He}, \bibinfo{person}{Sugato Basu}, \bibinfo{person}{Xin~Eric Wang}, {and} \bibinfo{person}{William~Yang Wang}.} \bibinfo{year}{2024}\natexlab{}.
\newblock \showarticletitle{Layoutgpt: Compositional visual planning and generation with large language models}.
\newblock \bibinfo{journal}{\emph{Advances in Neural Information Processing Systems}}  \bibinfo{volume}{36} (\bibinfo{year}{2024}).
\newblock


\bibitem[Fridman et~al\mbox{.}(2023)]%
        {SceneScape}
\bibfield{author}{\bibinfo{person}{Rafail Fridman}, \bibinfo{person}{Amit Abecasis}, \bibinfo{person}{Yoni Kasten}, {and} \bibinfo{person}{Tali Dekel}.} \bibinfo{year}{2023}\natexlab{}.
\newblock \showarticletitle{Scenescape: Text-driven consistent scene generation}.
\newblock \bibinfo{journal}{\emph{arXiv preprint arXiv:2302.01133}} (\bibinfo{year}{2023}).
\newblock


\bibitem[Fu et~al\mbox{.}(2025)]%
        {AnyHome}
\bibfield{author}{\bibinfo{person}{Rao Fu}, \bibinfo{person}{Zehao Wen}, \bibinfo{person}{Zichen Liu}, {and} \bibinfo{person}{Srinath Sridhar}.} \bibinfo{year}{2025}\natexlab{}.
\newblock \showarticletitle{Anyhome: Open-vocabulary generation of structured and textured 3d homes}. In \bibinfo{booktitle}{\emph{Proc. ECCV}}. Springer, \bibinfo{pages}{52--70}.
\newblock


\bibitem[Gasch et~al\mbox{.}(2022)]%
        {ecosystems2022}
\bibfield{author}{\bibinfo{person}{Cristina Gasch}, \bibinfo{person}{José Sotoca}, \bibinfo{person}{Miguel Chover}, \bibinfo{person}{Inmaculada Remolar}, {and} \bibinfo{person}{Cristina Rebollo}.} \bibinfo{year}{2022}\natexlab{}.
\newblock \showarticletitle{Procedural modeling of plant ecosystems maximizing vegetation cover}.
\newblock \bibinfo{journal}{\emph{Multimedia Tools and Applications}}  \bibinfo{volume}{81} (\bibinfo{date}{05} \bibinfo{year}{2022}).
\newblock
\urldef\tempurl%
\url{https://doi.org/10.1007/s11042-022-12107-8}
\showDOI{\tempurl}


\bibitem[Henzler et~al\mbox{.}(2019)]%
        {Henzler2019}
\bibfield{author}{\bibinfo{person}{Philipp Henzler}, \bibinfo{person}{Niloy~J Mitra}, {and} \bibinfo{person}{Tobias Ritschel}.} \bibinfo{year}{2019}\natexlab{}.
\newblock \showarticletitle{Escaping Plato's Cave: 3D shape from adversarial rendering}. In \bibinfo{booktitle}{\emph{Proceedings of the IEEE/CVF International Conference on Computer Vision}}. \bibinfo{pages}{9984--9993}.
\newblock


\bibitem[H{\"o}llein et~al\mbox{.}(2023)]%
        {Text2Room}
\bibfield{author}{\bibinfo{person}{Lukas H{\"o}llein}, \bibinfo{person}{Ang Cao}, \bibinfo{person}{Andrew Owens}, \bibinfo{person}{Justin Johnson}, {and} \bibinfo{person}{Matthias Nie{\ss}ner}.} \bibinfo{year}{2023}\natexlab{}.
\newblock \showarticletitle{Text2room: Extracting textured 3d meshes from 2d text-to-image models}. In \bibinfo{booktitle}{\emph{Proc. ICCV}}. \bibinfo{pages}{7909--7920}.
\newblock


\bibitem[Hsu et~al\mbox{.}(2023)]%
        {Ditto}
\bibfield{author}{\bibinfo{person}{Cheng-Chun Hsu}, \bibinfo{person}{Zhenyu Jiang}, {and} \bibinfo{person}{Yuke Zhu}.} \bibinfo{year}{2023}\natexlab{}.
\newblock \showarticletitle{Ditto in the house: Building articulation models of indoor scenes through interactive perception}. In \bibinfo{booktitle}{\emph{2023 IEEE International Conference on Robotics and Automation}}. IEEE, \bibinfo{pages}{3933--3939}.
\newblock


\bibitem[Hu et~al\mbox{.}(2024)]%
        {hu2024scenecraft}
\bibfield{author}{\bibinfo{person}{Ziniu Hu}, \bibinfo{person}{Ahmet Iscen}, \bibinfo{person}{Aashi Jain}, \bibinfo{person}{Thomas Kipf}, \bibinfo{person}{Yisong Yue}, \bibinfo{person}{David~A. Ross}, \bibinfo{person}{Cordelia Schmid}, {and} \bibinfo{person}{Alireza Fathi}.} \bibinfo{year}{2024}\natexlab{}.
\newblock \bibinfo{title}{SceneCraft: An LLM Agent for Synthesizing 3D Scene as Blender Code}.
\newblock
\newblock
\showeprint[arxiv]{2403.01248}~[cs.CV]


\bibitem[Lin and Yadong(2023)]%
        {InstructScene}
\bibfield{author}{\bibinfo{person}{Chenguo Lin} {and} \bibinfo{person}{MU Yadong}.} \bibinfo{year}{2023}\natexlab{}.
\newblock \showarticletitle{InstructScene: Instruction-Driven 3D Indoor Scene Synthesis with Semantic Graph Prior}. In \bibinfo{booktitle}{\emph{Proc. ICLR}}.
\newblock


\bibitem[Lin et~al\mbox{.}(2023)]%
        {Magic3D}
\bibfield{author}{\bibinfo{person}{Chen-Hsuan Lin}, \bibinfo{person}{Jun Gao}, \bibinfo{person}{Luming Tang}, \bibinfo{person}{Towaki Takikawa}, \bibinfo{person}{Xiaohui Zeng}, \bibinfo{person}{Xun Huang}, \bibinfo{person}{Karsten Kreis}, \bibinfo{person}{Sanja Fidler}, \bibinfo{person}{Ming-Yu Liu}, {and} \bibinfo{person}{Tsung-Yi Lin}.} \bibinfo{year}{2023}\natexlab{}.
\newblock \showarticletitle{Magic3d: High-resolution text-to-3d content creation}. In \bibinfo{booktitle}{\emph{Proceedings of the IEEE/CVF Conference on Computer Vision and Pattern Recognition}}. \bibinfo{pages}{300--309}.
\newblock


\bibitem[Lipp et~al\mbox{.}(2011)]%
        {lipp2011interactive}
\bibfield{author}{\bibinfo{person}{Markus Lipp}, \bibinfo{person}{Daniel Scherzer}, \bibinfo{person}{Peter Wonka}, {and} \bibinfo{person}{Michael Wimmer}.} \bibinfo{year}{2011}\natexlab{}.
\newblock \showarticletitle{Interactive modeling of city layouts using layers of procedural content}. In \bibinfo{booktitle}{\emph{Computer Graphics Forum}}, Vol.~\bibinfo{volume}{30}. Wiley Online Library, \bibinfo{pages}{345--354}.
\newblock


\bibitem[Liu et~al\mbox{.}(2023)]%
        {Zero123}
\bibfield{author}{\bibinfo{person}{Ruoshi Liu}, \bibinfo{person}{Rundi Wu}, \bibinfo{person}{Basile Van~Hoorick}, \bibinfo{person}{Pavel Tokmakov}, \bibinfo{person}{Sergey Zakharov}, {and} \bibinfo{person}{Carl Vondrick}.} \bibinfo{year}{2023}\natexlab{}.
\newblock \showarticletitle{Zero-1-to-3: Zero-shot one image to 3d object}. In \bibinfo{booktitle}{\emph{Proceedings of the IEEE/CVF International Conference on Computer Vision}}. \bibinfo{pages}{9298--9309}.
\newblock


\bibitem[Murali et~al\mbox{.}(2017)]%
        {Murali17}
\bibfield{author}{\bibinfo{person}{S. Murali}, \bibinfo{person}{P. Speciale}, \bibinfo{person}{M.~R. Oswald}, {and} \bibinfo{person}{M. Pollefeys}.} \bibinfo{year}{2017}\natexlab{}.
\newblock \showarticletitle{Indoor scan2bim: Building information models of house interiors}. In \bibinfo{booktitle}{\emph{2017 IEEE/RSJ International Conference on Intelligent Robots and Systems (IROS)}}. IEEE.
\newblock


\bibitem[Nguyen-Phuoc et~al\mbox{.}(2019)]%
        {NguyenPhuoc2019}
\bibfield{author}{\bibinfo{person}{Thu Nguyen-Phuoc}, \bibinfo{person}{Chuan Li}, \bibinfo{person}{Lucas Theis}, \bibinfo{person}{Christian Richardt}, {and} \bibinfo{person}{Yong-Liang Yang}.} \bibinfo{year}{2019}\natexlab{}.
\newblock \showarticletitle{Hologan: Unsupervised learning of 3D representations from natural images}. In \bibinfo{booktitle}{\emph{Proceedings of the IEEE/CVF International Conference on Computer Vision}}. \bibinfo{pages}{7588--7597}.
\newblock


\bibitem[Ochmann et~al\mbox{.}(2019)]%
        {Ochmann19}
\bibfield{author}{\bibinfo{person}{S. Ochmann}, \bibinfo{person}{R. Vock}, {and} \bibinfo{person}{R. Klein}.} \bibinfo{year}{2019}\natexlab{}.
\newblock \showarticletitle{Automatic reconstruction of fully volumetric 3d building models from oriented point clouds}.
\newblock \bibinfo{journal}{\emph{ISPRS Journal of Photogrammetry and Remote Sensing}}  \bibinfo{volume}{151} (\bibinfo{year}{2019}), \bibinfo{pages}{251--262}.
\newblock


\bibitem[Parish and Müller(2001)]%
        {PMC}
\bibfield{author}{\bibinfo{person}{Yoav Parish} {and} \bibinfo{person}{Pascal Müller}.} \bibinfo{year}{2001}\natexlab{}.
\newblock \showarticletitle{Procedural Modeling of Cities}.
\newblock \bibinfo{journal}{\emph{Proceedings of SIGGRAPH}}  \bibinfo{volume}{2001}, \bibinfo{pages}{301--308}.
\newblock
\urldef\tempurl%
\url{https://doi.org/10.1145/1185657.1185716}
\showDOI{\tempurl}


\bibitem[Poole et~al\mbox{.}(2022)]%
        {DreamFusion}
\bibfield{author}{\bibinfo{person}{Ben Poole}, \bibinfo{person}{Ajay Jain}, \bibinfo{person}{Jonathan~T Barron}, {and} \bibinfo{person}{Ben Mildenhall}.} \bibinfo{year}{2022}\natexlab{}.
\newblock \showarticletitle{DreamFusion: Text-to-3D using 2D Diffusion}. In \bibinfo{booktitle}{\emph{The Eleventh International Conference on Learning Representations}}.
\newblock


\bibitem[Raistrick et~al\mbox{.}(2023)]%
        {Infinigen}
\bibfield{author}{\bibinfo{person}{Alexander Raistrick}, \bibinfo{person}{Lahav Lipson}, \bibinfo{person}{Zeyu Ma}, \bibinfo{person}{Lingjie Mei}, \bibinfo{person}{Mingzhe Wang}, \bibinfo{person}{Yiming Zuo}, \bibinfo{person}{Karhan Kayan}, \bibinfo{person}{Hongyu Wen}, \bibinfo{person}{Beining Han}, \bibinfo{person}{Yihan Wang}, {et~al\mbox{.}}} \bibinfo{year}{2023}\natexlab{}.
\newblock \showarticletitle{Infinite Photorealistic Worlds using Procedural Generation}. In \bibinfo{booktitle}{\emph{Proceedings of the IEEE/CVF Conference on Computer Vision and Pattern Recognition}}. \bibinfo{pages}{12630--12641}.
\newblock


\bibitem[Raistrick et~al\mbox{.}(2024)]%
        {InfinigenIndoor}
\bibfield{author}{\bibinfo{person}{Alexander Raistrick}, \bibinfo{person}{Lingjie Mei}, \bibinfo{person}{Karhan Kayan}, \bibinfo{person}{David Yan}, \bibinfo{person}{Yiming Zuo}, \bibinfo{person}{Beining Han}, \bibinfo{person}{Hongyu Wen}, \bibinfo{person}{Meenal Parakh}, \bibinfo{person}{Stamatis Alexandropoulos}, \bibinfo{person}{Lahav Lipson}, {et~al\mbox{.}}} \bibinfo{year}{2024}\natexlab{}.
\newblock \showarticletitle{Infinigen Indoors: Photorealistic Indoor Scenes using Procedural Generation}. In \bibinfo{booktitle}{\emph{Proc. CVPR}}. \bibinfo{pages}{21783--21794}.
\newblock


\bibitem[Song et~al\mbox{.}(2023)]%
        {RoomDreamer}
\bibfield{author}{\bibinfo{person}{Liangchen Song}, \bibinfo{person}{Liangliang Cao}, \bibinfo{person}{Hongyu Xu}, \bibinfo{person}{Kai Kang}, \bibinfo{person}{Feng Tang}, \bibinfo{person}{Junsong Yuan}, {and} \bibinfo{person}{Yang Zhao}.} \bibinfo{year}{2023}\natexlab{}.
\newblock \showarticletitle{Roomdreamer: Text-driven 3d indoor scene synthesis with coherent geometry and texture}.
\newblock \bibinfo{journal}{\emph{arXiv preprint arXiv:2305.11337}} (\bibinfo{year}{2023}).
\newblock


\bibitem[Sun et~al\mbox{.}(2023)]%
        {3D-GPT}
\bibfield{author}{\bibinfo{person}{Chunyi Sun}, \bibinfo{person}{Junlin Han}, \bibinfo{person}{Weijian Deng}, \bibinfo{person}{Xinlong Wang}, \bibinfo{person}{Zishan Qin}, {and} \bibinfo{person}{Stephen Gould}.} \bibinfo{year}{2023}\natexlab{}.
\newblock \showarticletitle{3d-gpt: Procedural 3d modeling with large language models}.
\newblock \bibinfo{journal}{\emph{arXiv preprint arXiv:2310.12945}} (\bibinfo{year}{2023}).
\newblock


\bibitem[Talton et~al\mbox{.}(2011)]%
        {talton2011metropolis}
\bibfield{author}{\bibinfo{person}{Jerry~O Talton}, \bibinfo{person}{Yu Lou}, \bibinfo{person}{Steve Lesser}, \bibinfo{person}{Jared Duke}, \bibinfo{person}{Radom{\'\i}r Mech}, {and} \bibinfo{person}{Vladlen Koltun}.} \bibinfo{year}{2011}\natexlab{}.
\newblock \showarticletitle{Metropolis procedural modeling.}
\newblock \bibinfo{journal}{\emph{ACM Trans. Graph.}} \bibinfo{volume}{30}, \bibinfo{number}{2} (\bibinfo{year}{2011}), \bibinfo{pages}{11--1}.
\newblock


\bibitem[Tang et~al\mbox{.}(2024)]%
        {DiffuScene}
\bibfield{author}{\bibinfo{person}{Jiapeng Tang}, \bibinfo{person}{Yinyu Nie}, \bibinfo{person}{Lev Markhasin}, \bibinfo{person}{Angela Dai}, \bibinfo{person}{Justus Thies}, {and} \bibinfo{person}{Matthias Nie{\ss}ner}.} \bibinfo{year}{2024}\natexlab{}.
\newblock \showarticletitle{Diffuscene: Denoising diffusion models for generative indoor scene synthesis}. In \bibinfo{booktitle}{\emph{Proc. CVPR}}. \bibinfo{pages}{20507--20518}.
\newblock


\bibitem[Torne et~al\mbox{.}(2024)]%
        {torne2024reconciling}
\bibfield{author}{\bibinfo{person}{Marcel Torne}, \bibinfo{person}{Anthony Simeonov}, \bibinfo{person}{Zechu Li}, \bibinfo{person}{April Chan}, \bibinfo{person}{Tao Chen}, \bibinfo{person}{Abhishek Gupta}, {and} \bibinfo{person}{Pulkit Agrawal}.} \bibinfo{year}{2024}\natexlab{}.
\newblock \showarticletitle{Reconciling reality through simulation: A real-to-sim-to-real approach for robust manipulation}.
\newblock \bibinfo{journal}{\emph{arXiv preprint arXiv:2403.03949}} (\bibinfo{year}{2024}).
\newblock


\bibitem[Vanegas et~al\mbox{.}(2012)]%
        {Parcels}
\bibfield{author}{\bibinfo{person}{Carlos Vanegas}, \bibinfo{person}{Tom Kelly}, \bibinfo{person}{Basil Weber}, \bibinfo{person}{Jan Halatsch}, \bibinfo{person}{Daniel Aliaga}, {and} \bibinfo{person}{Pascal Müller}.} \bibinfo{year}{2012}\natexlab{}.
\newblock \showarticletitle{Procedural Generation of Parcels in Urban Modeling}.
\newblock \bibinfo{journal}{\emph{Computer Graphics Forum}}  \bibinfo{volume}{31} (\bibinfo{date}{05} \bibinfo{year}{2012}), \bibinfo{pages}{681--690}.
\newblock
\urldef\tempurl%
\url{https://doi.org/10.1111/j.1467-8659.2012.03047.x}
\showDOI{\tempurl}


\bibitem[Wang et~al\mbox{.}(2024)]%
        {Chat2Layout}
\bibfield{author}{\bibinfo{person}{Can Wang}, \bibinfo{person}{Hongliang Zhong}, \bibinfo{person}{Menglei Chai}, \bibinfo{person}{Mingming He}, \bibinfo{person}{Dongdong Chen}, {and} \bibinfo{person}{Jing Liao}.} \bibinfo{year}{2024}\natexlab{}.
\newblock \showarticletitle{Chat2Layout: Interactive 3D Furniture Layout with a Multimodal LLM}.
\newblock \bibinfo{journal}{\emph{arXiv preprint arXiv:2407.21333}} (\bibinfo{year}{2024}).
\newblock


\bibitem[Wu et~al\mbox{.}(2016)]%
        {Wu2016}
\bibfield{author}{\bibinfo{person}{Jiajun Wu}, \bibinfo{person}{Chengkai Zhang}, \bibinfo{person}{Tianfan Xue}, \bibinfo{person}{Bill Freeman}, {and} \bibinfo{person}{Josh Tenenbaum}.} \bibinfo{year}{2016}\natexlab{}.
\newblock \showarticletitle{Learning a probabilistic latent space of object shapes via 3D generative-adversarial modeling}. In \bibinfo{booktitle}{\emph{Advances in Neural Information Processing Systems}}, Vol.~\bibinfo{volume}{29}.
\newblock


\bibitem[Xie et~al\mbox{.}(2023)]%
        {xie2023citydreamer}
\bibfield{author}{\bibinfo{person}{Haozhe Xie}, \bibinfo{person}{Zhaoxi Chen}, \bibinfo{person}{Fangzhou Hong}, {and} \bibinfo{person}{Ziwei Liu}.} \bibinfo{year}{2023}\natexlab{}.
\newblock \showarticletitle{Citydreamer: Compositional generative model of unbounded 3d cities}.
\newblock \bibinfo{journal}{\emph{arXiv preprint arXiv:2309.00610}} (\bibinfo{year}{2023}).
\newblock


\bibitem[Yang et~al\mbox{.}(2019)]%
        {Yang2019}
\bibfield{author}{\bibinfo{person}{Guandao Yang}, \bibinfo{person}{Xun Huang}, \bibinfo{person}{Zekun Hao}, \bibinfo{person}{Ming-Yu Liu}, \bibinfo{person}{Serge Belongie}, {and} \bibinfo{person}{Bharath Hariharan}.} \bibinfo{year}{2019}\natexlab{}.
\newblock \showarticletitle{Pointflow: 3D point cloud generation with continuous normalizing flows}. In \bibinfo{booktitle}{\emph{Proceedings of the IEEE/CVF International Conference on Computer Vision}}. \bibinfo{pages}{4541--4550}.
\newblock


\bibitem[Yang et~al\mbox{.}(2024a)]%
        {LLplace}
\bibfield{author}{\bibinfo{person}{Yixuan Yang}, \bibinfo{person}{Junru Lu}, \bibinfo{person}{Zixiang Zhao}, \bibinfo{person}{Zhen Luo}, \bibinfo{person}{James~JQ Yu}, \bibinfo{person}{Victor Sanchez}, {and} \bibinfo{person}{Feng Zheng}.} \bibinfo{year}{2024}\natexlab{a}.
\newblock \showarticletitle{LLplace: The 3D Indoor Scene Layout Generation and Editing via Large Language Model}.
\newblock \bibinfo{journal}{\emph{arXiv preprint arXiv:2406.03866}} (\bibinfo{year}{2024}).
\newblock


\bibitem[Yang et~al\mbox{.}(2024b)]%
        {Holodeck}
\bibfield{author}{\bibinfo{person}{Yue Yang}, \bibinfo{person}{Fan-Yun Sun}, \bibinfo{person}{Luca Weihs}, \bibinfo{person}{Eli VanderBilt}, \bibinfo{person}{Alvaro Herrasti}, \bibinfo{person}{Winson Han}, \bibinfo{person}{Jiajun Wu}, \bibinfo{person}{Nick Haber}, \bibinfo{person}{Ranjay Krishna}, \bibinfo{person}{Lingjie Liu}, {et~al\mbox{.}}} \bibinfo{year}{2024}\natexlab{b}.
\newblock \showarticletitle{Holodeck: Language guided generation of 3d embodied ai environments}. In \bibinfo{booktitle}{\emph{Proc. CVPR}}. \bibinfo{pages}{16227--16237}.
\newblock


\bibitem[Yang et~al\mbox{.}(2013)]%
        {UrbanPattern2013}
\bibfield{author}{\bibinfo{person}{Yong-Liang Yang}, \bibinfo{person}{Jun Wang}, \bibinfo{person}{Etienne Vouga}, {and} \bibinfo{person}{Peter Wonka}.} \bibinfo{year}{2013}\natexlab{}.
\newblock \showarticletitle{Urban Pattern: Layout Design by Hierarchical Domain Splitting}.
\newblock \bibinfo{journal}{\emph{ACM Transactions on Graphics (Proceedings of SIGGRAPH Asia 2013)}}  \bibinfo{volume}{32} (\bibinfo{year}{2013}), \bibinfo{pages}{Article No. xx}.
\newblock
Issue 6.


\bibitem[Zhai et~al\mbox{.}(2025)]%
        {EchoScene}
\bibfield{author}{\bibinfo{person}{Guangyao Zhai}, \bibinfo{person}{Evin~P{\i}nar {\"O}rnek}, \bibinfo{person}{Dave~Zhenyu Chen}, \bibinfo{person}{Ruotong Liao}, \bibinfo{person}{Yan Di}, \bibinfo{person}{Nassir Navab}, \bibinfo{person}{Federico Tombari}, {and} \bibinfo{person}{Benjamin Busam}.} \bibinfo{year}{2025}\natexlab{}.
\newblock \showarticletitle{Echoscene: Indoor scene generation via information echo over scene graph diffusion}. In \bibinfo{booktitle}{\emph{Proc. ECCV}}. Springer, \bibinfo{pages}{167--184}.
\newblock


\bibitem[Zhang et~al\mbox{.}(2024a)]%
        {FurniScene}
\bibfield{author}{\bibinfo{person}{Genghao Zhang}, \bibinfo{person}{Yuxi Wang}, \bibinfo{person}{Chuanchen Luo}, \bibinfo{person}{Shibiao Xu}, \bibinfo{person}{Junran Peng}, \bibinfo{person}{Zhaoxiang Zhang}, {and} \bibinfo{person}{Man Zhang}.} \bibinfo{year}{2024}\natexlab{a}.
\newblock \showarticletitle{FurniScene: A Large-scale 3D Room Dataset with Intricate Furnishing Scenes}.
\newblock \bibinfo{journal}{\emph{arXiv preprint arXiv:2401.03470}} (\bibinfo{year}{2024}).
\newblock


\bibitem[Zhang et~al\mbox{.}(2023a)]%
        {Text2Nerf}
\bibfield{author}{\bibinfo{person}{Jingbo Zhang}, \bibinfo{person}{Xiaoyu Li}, \bibinfo{person}{Ziyu Wan}, \bibinfo{person}{Can Wang}, {and} \bibinfo{person}{Jing Liao}.} \bibinfo{year}{2023}\natexlab{a}.
\newblock \showarticletitle{Text2NeRF: Text-Driven 3D Scene Generation with Neural Radiance Fields}.
\newblock \bibinfo{journal}{\emph{arXiv preprint arXiv:2305.11588}} (\bibinfo{year}{2023}).
\newblock


\bibitem[Zhang et~al\mbox{.}(2019)]%
        {rivers}
\bibfield{author}{\bibinfo{person}{Jian Zhang}, \bibinfo{person}{Chang-bo Wang}, \bibinfo{person}{Hong Qin}, \bibinfo{person}{Yi Chen}, {and} \bibinfo{person}{Yan Gao}.} \bibinfo{year}{2019}\natexlab{}.
\newblock \showarticletitle{Procedural modeling of rivers from single image toward natural scene production}.
\newblock \bibinfo{journal}{\emph{The Visual Computer}}  \bibinfo{volume}{35} (\bibinfo{date}{02} \bibinfo{year}{2019}).
\newblock
\urldef\tempurl%
\url{https://doi.org/10.1007/s00371-017-1465-7}
\showDOI{\tempurl}


\bibitem[Zhang et~al\mbox{.}(2023b)]%
        {SceneWiz3D}
\bibfield{author}{\bibinfo{person}{Qihang Zhang}, \bibinfo{person}{Chaoyang Wang}, \bibinfo{person}{Aliaksandr Siarohin}, \bibinfo{person}{Peiye Zhuang}, \bibinfo{person}{Yinghao Xu}, \bibinfo{person}{Ceyuan Yang}, \bibinfo{person}{Dahua Lin}, \bibinfo{person}{Bolei Zhou}, \bibinfo{person}{Sergey Tulyakov}, {and} \bibinfo{person}{Hsin-Ying Lee}.} \bibinfo{year}{2023}\natexlab{b}.
\newblock \showarticletitle{SceneWiz3D: Towards Text-guided 3D Scene Composition}.
\newblock \bibinfo{journal}{\emph{arXiv preprint arXiv:2312.08885}} (\bibinfo{year}{2023}).
\newblock


\bibitem[Zhang et~al\mbox{.}(2024b)]%
        {cityX}
\bibfield{author}{\bibinfo{person}{Shougao Zhang}, \bibinfo{person}{Mengqi Zhou}, \bibinfo{person}{Yuxi Wang}, \bibinfo{person}{Chuanchen Luo}, \bibinfo{person}{Rongyu Wang}, \bibinfo{person}{Yiwei Li}, \bibinfo{person}{Xucheng Yin}, \bibinfo{person}{Zhaoxiang Zhang}, {and} \bibinfo{person}{Junran Peng}.} \bibinfo{year}{2024}\natexlab{b}.
\newblock \showarticletitle{CityX: Controllable Procedural Content Generation for Unbounded 3D Cities}.
\newblock \bibinfo{journal}{\emph{arXiv preprint arXiv:2407.17572}} (\bibinfo{year}{2024}).
\newblock


\bibitem[Zhou et~al\mbox{.}(2021)]%
        {Zhou2021}
\bibfield{author}{\bibinfo{person}{Linqi Zhou}, \bibinfo{person}{Yilun Du}, {and} \bibinfo{person}{Jiajun Wu}.} \bibinfo{year}{2021}\natexlab{}.
\newblock \showarticletitle{3D shape generation and completion through point-voxel diffusion}. In \bibinfo{booktitle}{\emph{Proceedings of the IEEE/CVF International Conference on Computer Vision}}. \bibinfo{pages}{5826--5835}.
\newblock


\bibitem[Zhou et~al\mbox{.}(2024)]%
        {SceneX}
\bibfield{author}{\bibinfo{person}{Mengqi Zhou}, \bibinfo{person}{Yuxi Wang}, \bibinfo{person}{Jun Hou}, \bibinfo{person}{Chuanchen Luo}, \bibinfo{person}{Zhaoxiang Zhang}, {and} \bibinfo{person}{Junran Peng}.} \bibinfo{year}{2024}\natexlab{}.
\newblock \showarticletitle{SceneX: Procedural controllable large-scale scene generation via large-language models}.
\newblock \bibinfo{journal}{\emph{arXiv preprint arXiv:2403.15698}} (\bibinfo{year}{2024}).
\newblock


\end{thebibliography}
